\documentclass[11pt,a4paper,copyright,nonumbering]{mll_style}
\usepackage[numbers,sort&compress]{natbib}
\usepackage{nicefrac}
\usepackage{enumitem}
\usepackage{subcaption}
\usepackage{multirow}
\usepackage{wrapfig}
\usepackage{placeins}
\usepackage{float}
\usepackage[most]{tcolorbox}
\newcommand{\method}{DepthGraft}
\reportnumber{001}
\fancypagestyle{depthgraftfirst}[firststyle]{
    \fancyfoot[L,R]{}
    \fancyfoot[C]{%
        \parbox[t]{\textwidth}{\raggedright
            \normalfont\fontsize{8.5}{10}\selectfont
            $^\dagger$Corresponding author:
            \href{mailto:yanjunchi@sjtu.edu.cn}{\nolinkurl{yanjunchi@sjtu.edu.cn}}.\\[3pt]
            \makebox[\textwidth][c]{\footerfont\thepage}%
        }%
    }
}
\title{\centering DepthGraft: Structural Regularization through
Hierarchical Cross-Layer KV Reconstruction}
\date{}
\author[*]{%
Xiaohan Qin$^{1,2,3}$,
Xiangdong Zhang$^{1,3}$,
Yu Wang$^{3}$,
Huaijin Wu$^{1,3}$,\authorcr
Zhuo Xia$^{1,3}$,
Yebin Yang$^{1,3}$,
Junchi Yan$^{1,\dagger}$\\
{\small $^1$School of AI, Shanghai Jiao Tong University~~~$^2$Shanghai Innovation Institute}\\
{\small $^3$Dots Studio, Xiaohongshu Inc.}\\
{\normalfont\small \textbf{Project page:}
\href{https://github.com/jianke0604/DepthGraft}{\nolinkurl{https://github.com/jianke0604/DepthGraft}}}
}

\begin{abstract}
Scaling depth is a key route to stronger large language models, but Pre-LN architectures often suffer from depth-wise information dilution, limiting the effective use of deep layers.
While cross-layer connectivity has proven effective at mitigating this problem, existing designs usually introduce additional computation or memory access.
We revisit cross-layer KV sharing as an efficient form of such connectivity.
Beyond cache compression, each KV reuse edge creates a direct gradient path from deep layers to source KV projections, turning KV sharing into an implicit structural regularizer for KV representation geometry.
We provide a systematic analysis of this mechanism, showing that proper sharing topologies sharpen key retrieval directions while enriching the value content subspace, whereas existing adjacent or global-anchor designs do not fully exploit this effect.
Motivated by this, we propose DepthGraft, a hierarchical KV-sharing architecture that reconstructs deep-layer KV from stride-aligned shallow--middle source pairs. DepthGraft distributes non-local gradient flow across source layers, while preserving prefill early completion and reducing KV cache by roughly one third.
Experiments on dense and MoE models from 0.5B to 16B parameters show consistent downstream improvements under both GQA and MLA.
Notably, on the 16B MoE model, \method{} achieves a \textbf{7.0\%} absolute improvement on MMLU-Pro, along with gains of \textbf{6.6\%} on CMMLU and \textbf{6.4\%} on C3.
These results validate the potential of \method{} as a scalable design for large-scale pretraining.
The project page is available at \url{https://github.com/jianke0604/DepthGraft}.

\end{abstract}

\begin{document}
\maketitle
\thispagestyle{depthgraftfirst}

\begin{figure}[!h]
    \centering
    \begin{subfigure}[b]{0.24\textwidth}
        \includegraphics[width=\textwidth]{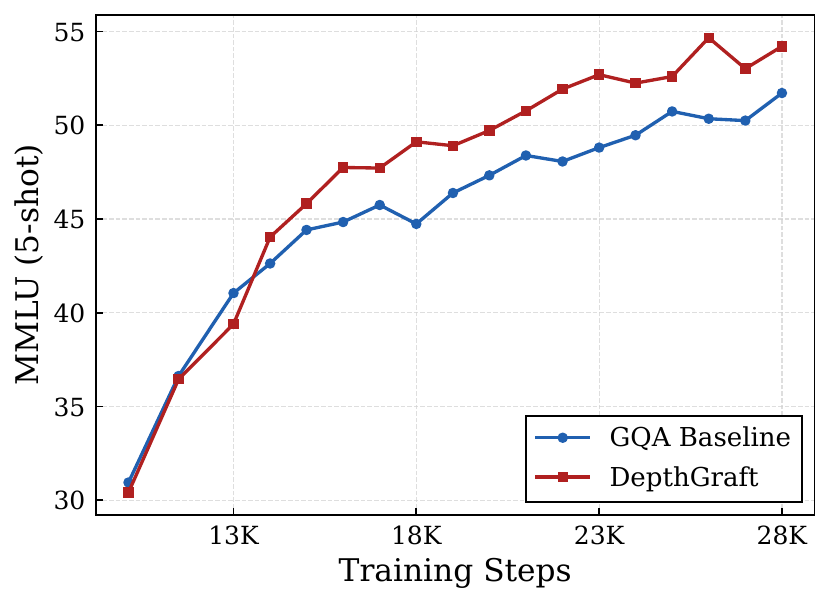}
    \end{subfigure}
    \hfill
    \begin{subfigure}[b]{0.24\textwidth}
        \includegraphics[width=\textwidth]{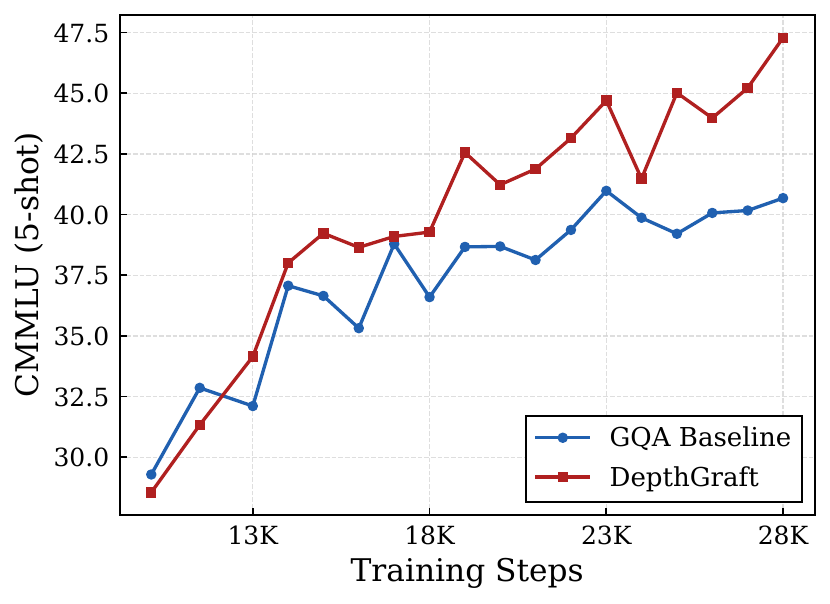}
    \end{subfigure}
    \hfill
    \begin{subfigure}[b]{0.24\textwidth}
        \includegraphics[width=\textwidth]{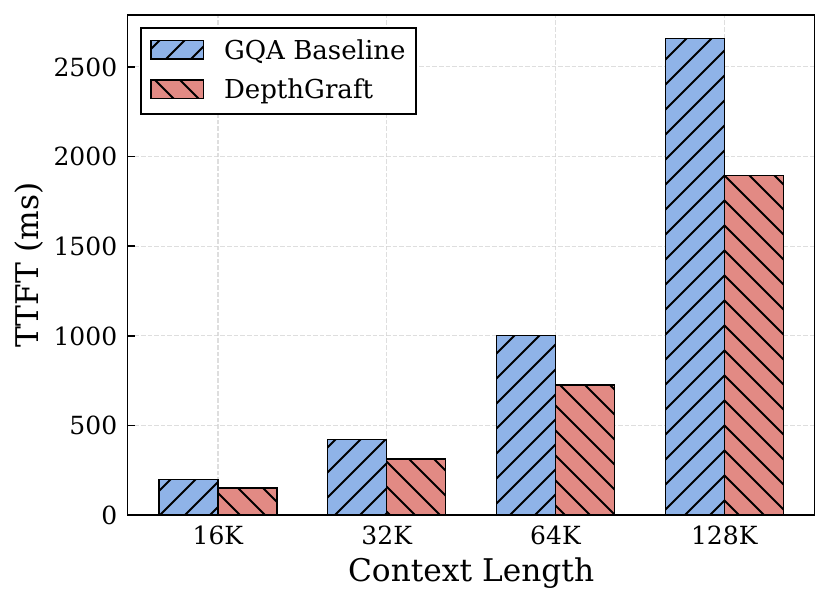}
    \end{subfigure}
    \hfill
    \begin{subfigure}[b]{0.24\textwidth}
        \includegraphics[width=\textwidth]{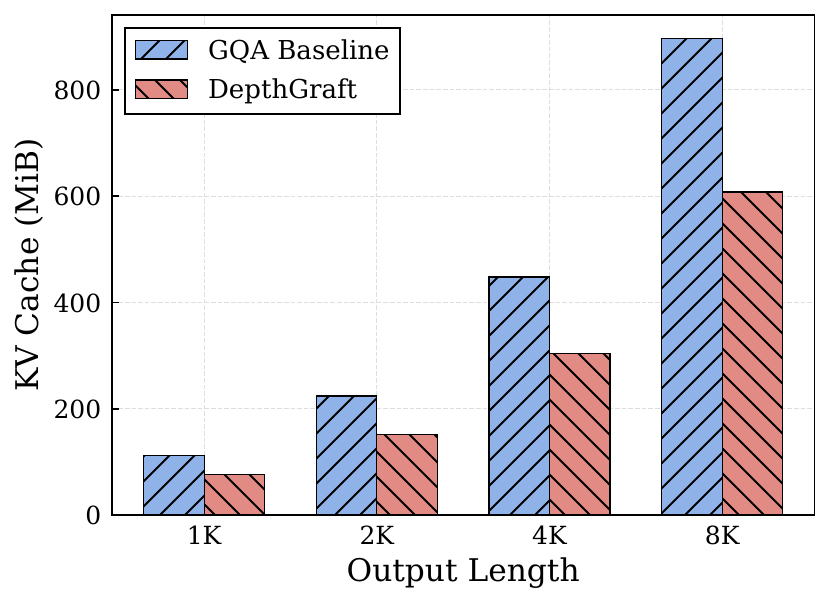}
    \end{subfigure}
    \caption{Results on the 16B MoE model. \textbf{Left two:} DepthGraft demonstrates a consistent advantage throughout training. \textbf{Right two:} DepthGraft reduces prefill time (TTFT) and KV cache usage by approximately one-third, while incurring almost no loss in decoding speed.}
    \label{fig:benchmark_curves}
\end{figure}

\section{Introduction}
\label{sec:intro}

Scaling depth has been a primary lever for improving Transformer-based large language models (LLMs)~\cite{yang2025qwen3,team2025kimi,liu2025deepseek}.
In modern Pre-Layer-Normalization Transformers~\cite{xiong2020layer}, however, adding layers does not always yield the expected improvements~\cite{gromov2024unreasonable,men2025shortgpt}.
The residual stream~\cite{he2016deep} aggregates all preceding computations through a single hidden-state trajectory with fixed additive updates; as the stack grows deeper, features formed in early layers can be progressively diluted and become difficult for later layers to retrieve.
This depth-wise information dilution limits the effective use of additional layers, making depth scaling less efficient than its representational capacity would suggest~\citep{sun2025curse,csordas2025language}.

Recent work has validated \emph{cross-layer connectivity} as an effective remedy for information dilution along depth.
Instead of forcing all depth history through the residual stream, recent designs allow deep layers to directly read earlier representations: Attention Residuals~\citep{team2026attention} reweights preceding hidden states to replace fixed residual accumulation, while Mixture-of-Depths Attention (MoDA)~\citep{zhu2026mixture} augments attention with key--value pairs from previous layers.
Such connections give deep layers direct access to earlier features and create shorter gradient paths to shallow layers.
However, dense cross-layer access also introduces additional computation, memory traffic, and system complexity, requiring careful performance--efficiency trade-offs and sophisticated systems-level engineering.

We note that cross-layer KV sharing can also be viewed as a form of such connectivity:
it provides deep layers with a channel that \emph{bypasses} information dilution in the residual stream by granting direct access to shallow-layer memory representations,
while gradients flow from deep layers back through the shared KV to shallow layers,
supplying shallow-layer KV with optimization signals driven by deep-layer task demands---thus holding the potential to improve depth utilization.
Although cross-layer KV sharing has been explored in prior work~\cite{brandon2024reducing,zuhri2025mlkv,sun2024you}, existing methods predominantly approach it from the perspective of cache compression and inference acceleration.
No prior work has systematically studied the cross-layer gradient dynamics of KV sharing, especially how such connections reshape KV representation geometry during pretraining and how the sharing topology determines this effect.

In this paper, we propose that KV sharing is not merely a cache-saving mechanism, but also a form of \emph{cross-layer gradient regularization}.
When a source layer's KV is reused by a distant deep layer, the source KV receives training signals from multiple query spaces.
This reshapes KV representation geometry in an asymmetric way: keys are encouraged to concentrate around shared discriminative retrieval directions, while values are pushed to cover a richer content subspace.
With a proper sharing topology, this specialization sharpens memory retrieval while increasing the expressive capacity of the value content channel, yielding a useful structural bias for training.
We use this lens to revisit existing KV-sharing methods and find that their topologies expose complementary limitations.
Adjacent-layer sharing provides too little query-space separation, whereas global-anchor sharing concentrates heterogeneous deep-layer gradients on a few sources, preventing either design from fully exploiting the regularization effect.

Motivated by this analysis, we propose \textbf{DepthGraft}, a simple hierarchical KV-sharing design derived from the cross-layer regularization view.
DepthGraft partitions the network into shallow, middle, and deep stages.
The shallow and middle stages compute KV normally, while each deep layer reconstructs its KV from a stride-aligned shallow--middle source pair using learnable channel-wise coefficients.
This construction creates non-local source--reuse connections with sufficient query-space separation, distributes cross-layer gradients across many source layers rather than a few global anchors, and gives each deep layer access to memory bases at two abstraction levels.
As a result, DepthGraft preserves the efficiency benefits of KV sharing, including prefill early completion and KV-cache reduction, while turning the memory pathway into an effective regularized cross-layer connection.
Training-dynamics analyses further support our theory, showing that DepthGraft reshapes KV representation geometry as predicted, and experiments across dense and MoE models demonstrate consistent performance gains at multiple scales.

Our main contributions are as follows.

\textbf{1). KV sharing as cross-layer gradient regularization.}
We identify KV sharing as a form of cross-layer connectivity and provide a systematic analysis of how its gradient paths reshape KV geometry.
Our analysis shows that proper KV sharing sharpens key retrieval directions while enriching the value content subspace, and that this effect is governed by the sharing topology.

\textbf{2). Hierarchical cross-layer KV reconstruction.}
We propose \method{}, which uses a hierarchical stage layout to reconstruct deep-layer KV from stride-aligned shallow--middle source pairs.
This yields a distributed non-local connectivity pattern with two abstraction-level memory bases, avoiding the concentrated-anchor structure of prior KV-sharing methods.

\textbf{3). Efficient and scalable implementation.}
\method{} removes independent KV materialization from the deep stage, reducing KV cache by roughly $\frac{1}{3}$ and allowing KV construction to finish before the final stage during prefill.
Its stride-aligned source assignment is engineering-friendly and naturally compatible with PP/VPP schedules, making the design suitable for large-scale distributed training.

\textbf{4). Consistent empirical gains.}
We validate \method{} on both dense and MoE models, covering model scales from 0.5B to 16B total parameters.
Across scales and attention variants, \method{} consistently improves downstream performance while preserving inference efficiency.

\section{Related Work}
\label{sec:related}

\textbf{Cross-layer connectivity.}
A broad line of work enriches depth-wise information flow by giving each layer direct access to earlier representations~\cite{huang2017densely,ilic2018deep,knupp2026depth}.
DenseNet~\citep{huang2017densely} concatenates all preceding feature maps, establishing a canonical dense cross-layer design.
In Transformers, DenseFormer~\citep{pagliardini2024denseformer} and ANCRe~\citep{zhang2026ancre} introduce learned layer-wise connection weights, while Value Residual Learning~\citep{zhou2025value} adds targeted residual access to earlier value states.
Attention Residuals~\citep{team2026attention} replaces fixed residual accumulation with learned reweighting over preceding hidden states, and MoDA~\citep{zhu2026mixture} augments sequence attention with key--value memories from previous layers.
Together, these works show that cross-layer connectivity is an effective mechanism for improving deep Transformer architectures~\citep{zhang2026xhc,qin2026multi}.

\textbf{KV sharing for efficient inference.}
KV cache is a major memory bottleneck during autoregressive inference, motivating many methods that reduce cache storage through KV sharing~\citep{wu2025hshare,yang2024kvsharer,wu2024layer}.
Cross-layer KV sharing is an important direction in this line.
CLA~\citep{brandon2024reducing} and MLKV~\citep{zuhri2025mlkv} reuse KV within adjacent layer groups, reducing cache but often incurring performance degradation when the reuse constraint becomes too strong.
YOCO~\citep{sun2024you} shares a global KV memory for all deeper layers, enabling prefill early completion and strong cache reduction.
FusedKV~\citep{lin2025reconstructing} further reconstructs deep-layer KV from two global anchors, providing more flexibility than direct global reuse.
Cross-layer KV sharing is becoming an increasingly important architectural choice for large-scale LLM pretraining, as illustrated by its adoption in DeepSeek-V4.1-Flash~\citep{xu2026deepseek} and HySparse2~\citep{wei2026hysparse2}.

\textbf{Positioning.}
\method{} systematically analyzes how cross-layer KV sharing alters training dynamics, showing that the induced cross-layer gradients can regularize KV representation geometry.
Guided by this analysis, \method{} uses stride-aligned hierarchical reconstruction to improve KV representations and model quality while preserving the KV-cache savings of sharing-based inference.

\section{Method}
\label{sec:method}

\subsection{Preliminaries}
\label{sec:prelim}

\textbf{Attention and KV cache.}
Consider an $L$-layer decoder-only Transformer. We index layers by
$\ell \in \{0,\ldots,L-1\}$ and use hidden states
$\mathbf{X}_\ell \in \mathbb{R}^{T \times D}$ for a sequence of length $T$.
Layer $\ell$ forms queries, keys, and values as
\begin{equation}
    \mathbf{Q}_\ell = \mathbf{X}_\ell \mathbf{W}_\ell^Q, \qquad
    \mathbf{K}_\ell = \mathbf{X}_\ell \mathbf{W}_\ell^K, \qquad
    \mathbf{V}_\ell = \mathbf{X}_\ell \mathbf{W}_\ell^V,
    \label{eq:qkv}
\end{equation}
where $\mathbf{W}_\ell^Q \in \mathbb{R}^{D \times H_q d}$, $\mathbf{W}_\ell^K, \mathbf{W}_\ell^V \in \mathbb{R}^{D \times H_k d}$,
with $H_q$ query heads, $H_k$ key-value heads, and per-head dimension $d$.
Under grouped-query attention~\citep{ainslie2023gqa}, the $H_q$ query heads are assigned to $H_k$ KV groups.
For query head $h\in\{0,\ldots,H_q-1\}$, let $\phi(h)=\lfloor h H_k/H_q \rfloor$ denote its KV group. The attention output is
\begin{equation}
    \operatorname{Attn}_\ell = \operatorname{Concat}_{h=0}^{H_q-1}\!\left(
        \operatorname{softmax}\!\Bigl(\frac{\mathbf{Q}_{\ell,h}\, \mathbf{K}_{\ell,\phi(h)}^\top}{\sqrt{d}}\Bigr)\, \mathbf{V}_{\ell,\phi(h)}
    \right) \mathbf{W}_\ell^O,
    \label{eq:attn}
\end{equation}
where $\mathbf{Q}_{\ell,h},\mathbf{K}_{\ell,g},\mathbf{V}_{\ell,g}\in\mathbb{R}^{T\times d}$ and
$\mathbf{W}_\ell^O \in \mathbb{R}^{H_q d \times D}$.
In a standard Transformer, each layer materializes its own $(\mathbf{K}_\ell,\mathbf{V}_\ell)$, so autoregressive inference caches the full set
$\{(\mathbf{K}_\ell,\mathbf{V}_\ell)\}_{\ell=0}^{L-1}$.

\textbf{Cross-layer KV sharing as memory routing.}
KV sharing replaces some of these independent KV tensors with KV tensors produced at other layers.
We write $(\mathbf{K}_\ell^{\mathrm{use}},\mathbf{V}_\ell^{\mathrm{use}})$ for the pair actually used by layer $\ell$ in Eq.~\ref{eq:attn}.
This pair may be computed locally, $(\mathbf{K}_\ell^{\mathrm{use}},\mathbf{V}_\ell^{\mathrm{use}})=(\mathbf{K}_\ell,\mathbf{V}_\ell)$, or derived from a set of source layers.
Thus a KV-sharing scheme specifies a cross-layer memory-routing graph: source layers produce memory states, and reuse layers consume them.
This view separates three design choices that matter for both efficiency and optimization: which layers serve as sources, how many reuse layers each source serves, and how the reused KV is constructed.

\emph{Adjacent-layer sharing.}
CLA~\citep{brandon2024reducing} and MLKV~\citep{zuhri2025mlkv} partition consecutive layers into groups of size $N$.
Within each group, only the first layer computes KV and the remaining layers reuse it:
\begin{equation}
    \mathbf{K}_\ell^{\mathrm{use}} = \mathbf{K}_{\lfloor \ell / N \rfloor \cdot N},
    \qquad
    \mathbf{V}_\ell^{\mathrm{use}} = \mathbf{V}_{\lfloor \ell / N \rfloor \cdot N}.
    \label{eq:cla}
\end{equation}
Because source layers are interleaved throughout depth, this pattern reduces the number of cached KV tensors but does not make the cache complete before the final layer group is processed.

\emph{Global-anchor sharing.}
Global-anchor methods place all reusable memory before a boundary layer $n$.
For layers $\ell>n$, KV is no longer computed independently but obtained from one or more anchors.
YOCO~\citep{sun2024you} uses a single anchor:
$\mathbf{K}_\ell^{\mathrm{use}} = \mathbf{K}_n,\; \mathbf{V}_\ell^{\mathrm{use}} = \mathbf{V}_n,\; \forall\, \ell > n$.
FusedKV~\citep{lin2025reconstructing} instead reconstructs deep-layer KV from two anchors through channel-wise fusion:
\begin{equation}
    \mathbf{K}_\ell^{\mathrm{use}} = \mathcal{F}_\ell^K(\mathbf{K}_0, \mathbf{K}_n), \qquad \mathbf{V}_\ell^{\mathrm{use}} = \mathcal{F}_\ell^V(\mathbf{V}_0, \mathbf{V}_n),
    \qquad \forall\, \ell > n,
    \label{eq:fusedkv}
\end{equation}
where $\mathcal{F}_\ell^K$ and $\mathcal{F}_\ell^V$ are per-layer fusion functions.
Since all source KV tensors are available once layer $n$ completes, global-anchor designs already have all the KV states needed by the remaining layers, thereby enabling prefill early completion.

\textbf{Forward reuse induces backward connectivity.}
Although KV sharing is usually introduced as a cache-saving mechanism, it also changes the training dynamics.
If layer $\ell$ uses KV derived from a source layer $s$, gradients from layer $\ell$ backpropagate through the reused tensors to the source projections $\mathbf{W}_s^K$ and $\mathbf{W}_s^V$.
Thus every memory-routing edge is also a direct cross-layer gradient path.
We next analyze how these cross-layer gradient signals change the representation geometry of keys and values during training.

\subsection{Representation Geometry of Keys and Values Under Cross-Layer Sharing}
\label{sec:analysis}

We use \emph{effective rank}~\citep{roy2007effective} as a compact and widely used measure of representation dimensionality.
For a centered representation matrix $\mathbf{M}\in\mathbb{R}^{N\times d}$, we compute its feature covariance $\mathbf{C}_{\mathbf{M}}=\mathbf{M}^\top \mathbf{M}/N$ with eigenvalues $\{\lambda_i\}_{i=1}^{d}$, and define
\begin{equation}
    \operatorname{erank}(\mathbf{C}_{\mathbf{M}}) = \exp\!\Bigl(-\sum_{i=1}^{d} p_i \log p_i\Bigr),
    \qquad p_i = \frac{\lambda_i}{\sum_{j=1}^{d}\lambda_j}.
    \label{eq:erank}
\end{equation}
Applied to $\mathbf{K}_\ell$ and $\mathbf{V}_\ell$, effective rank measures how evenly activation energy is distributed across feature directions, reflecting the dimensionality of the key retrieval geometry and value content space.

\begin{figure}[t]
    \centering
    \begin{subfigure}[b]{0.24\textwidth}
        \includegraphics[width=\textwidth]{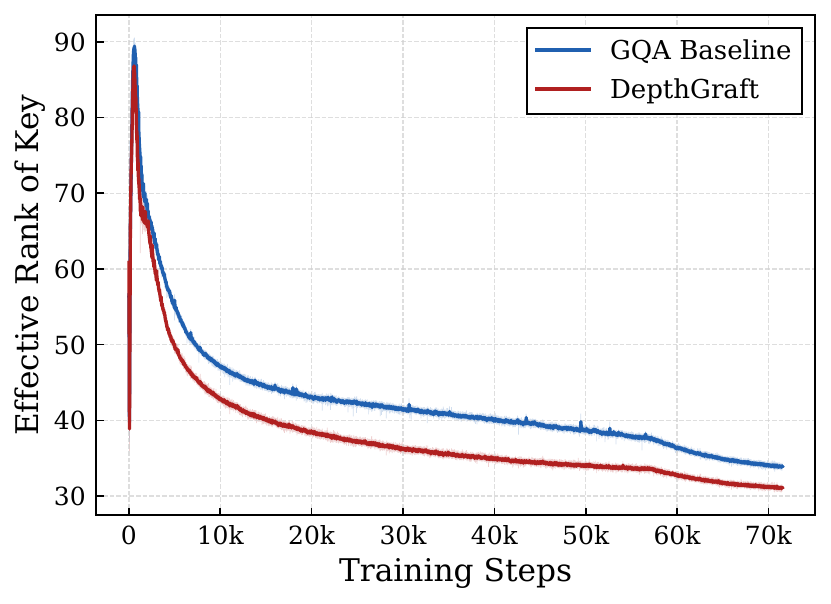}
    \end{subfigure}
    \hfill
    \begin{subfigure}[b]{0.24\textwidth}
        \includegraphics[width=\textwidth]{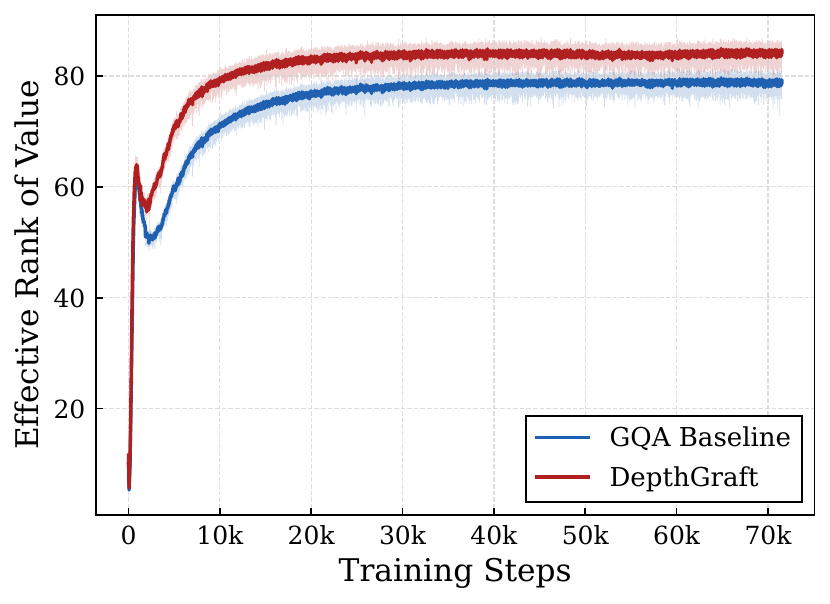}
    \end{subfigure}
    \hfill
    \begin{subfigure}[b]{0.24\textwidth}
        \includegraphics[width=\textwidth]{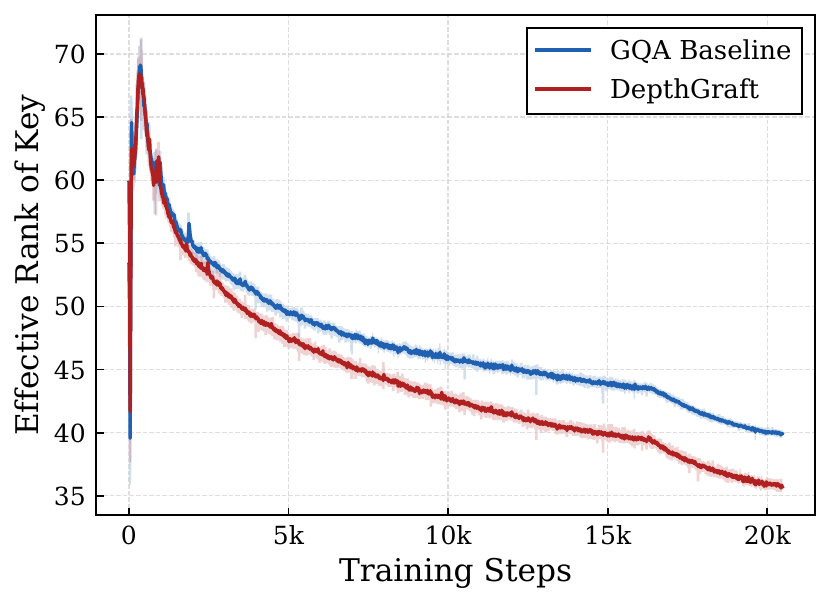}
    \end{subfigure}
    \hfill
    \begin{subfigure}[b]{0.24\textwidth}
        \includegraphics[width=\textwidth]{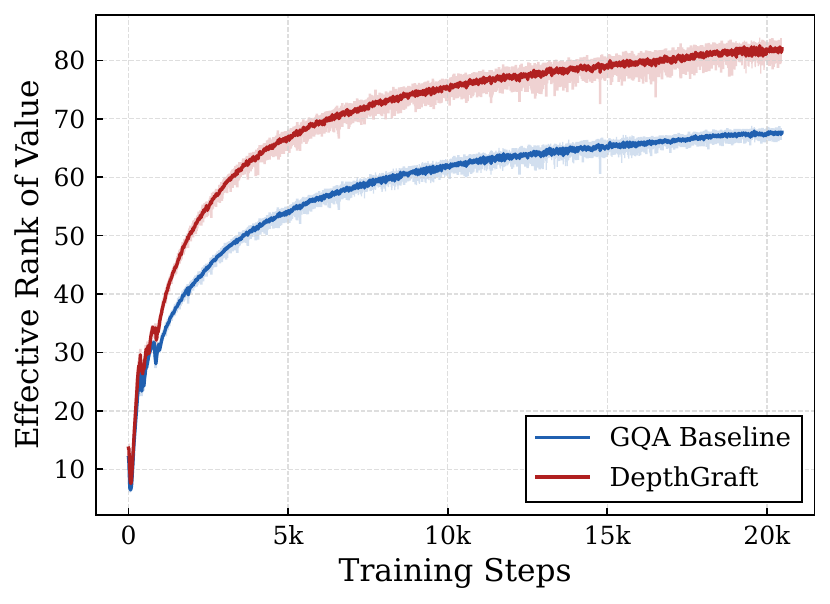}
    \end{subfigure}
    \\[-2pt]
    \makebox[0.48\textwidth]{\small (a) 1.5B Dense}
    \hfill
    \makebox[0.48\textwidth]{\small (b) 9B MoE}
    \caption{Evolution of KV representation geometry during training, visualized via effective rank (averaged over all layers).}
    \label{fig:kv_erank}
\end{figure}

\textbf{K/V geometry in standard attention.}
Suppressing head indices, let
$\mathbf{S}_\ell = \mathbf{Q}_\ell \mathbf{K}_\ell^\top/\sqrt{d}$,
$\mathbf{P}_\ell=\operatorname{softmax}(\mathbf{S}_\ell)$, $\mathbf{O}_\ell=\mathbf{P}_\ell \mathbf{V}_\ell$,
$\mathbf{G}_\ell=\partial\mathcal{L}/\partial \mathbf{S}_\ell$, and $\mathbf{E}_\ell=\partial\mathcal{L}/\partial \mathbf{O}_\ell$.
In a standard layer,
\begin{equation}
    \nabla_{\mathbf{K}_\ell} = \tfrac{1}{\sqrt{d}}\, \mathbf{G}_\ell^\top \mathbf{Q}_\ell,
    \qquad
    \nabla_{\mathbf{V}_\ell} = \mathbf{P}_\ell^\top \mathbf{E}_\ell.
    \label{eq:grad_std}
\end{equation}
These two expressions expose the asymmetry between K and V.
Keys are optimized through the softmax logits, where the competitive normalization makes retrieval decisions depend on relative token scores.
During training, key directions that repeatedly create useful logit separations receive stronger reinforcement, while directions that do not affect the induced attention distribution are less consistently updated.
This produces a natural tendency for the key spectrum to concentrate along a smaller set of discriminative directions.
As shown by the baseline curves in Fig.~\ref{fig:kv_erank}, the effective rank of $\mathbf{K}_\ell$ decreases during training, reflecting the formation of a sharper retrieval geometry.

Values enter attention through a different pathway.
Once $\mathbf{P}_\ell$ is fixed, the output is a linear readout $\mathbf{O}_\ell=\mathbf{P}_\ell \mathbf{V}_\ell$.
The value gradient $\mathbf{P}_\ell^\top \mathbf{E}_\ell$ routes downstream output gradients back to the value matrix according to the current attention pattern.
As the model learns to transmit richer information through attention, this linear content channel is encouraged to use more independent directions.
Correspondingly, the baseline in Fig.~\ref{fig:kv_erank} shows that the effective rank of $\mathbf{V}_\ell$ increases during training.

\textbf{KV sharing amplifies K/V specialization through cross-layer gradients.}
Consider a source layer $s$ whose KV is reused by a distant layer $\ell$.
The source KV receives both its local gradient and the reuse-layer gradient:
\begin{equation}
    \nabla_{\mathbf{K}_s}^{\mathrm{share}} = \tfrac{1}{\sqrt{d}}\, \mathbf{G}_s^\top \mathbf{Q}_s + \tfrac{1}{\sqrt{d}}\, \mathbf{G}_\ell^\top \mathbf{Q}_\ell,
    \qquad
    \nabla_{\mathbf{V}_s}^{\mathrm{share}} = \mathbf{P}_s^\top \mathbf{E}_s + \mathbf{P}_\ell^\top \mathbf{E}_\ell.
    \label{eq:grad_share}
\end{equation}
The same source key now serves two query spaces, and the same source value supports two readout patterns.
When $s$ and $\ell$ are sufficiently separated, $\mathbf{Q}_s$ and $\mathbf{Q}_\ell$ share task-relevant retrieval directions but also contain layer-specific components.
For keys, shared discriminative directions receive coherent reinforcement, whereas layer-specific directions can interfere through the softmax competition; the source key is therefore biased toward directions useful across both layers, further sharpening the retrieval geometry.
For values, the two readout terms do not compete through softmax.
Different attention patterns instead route complementary gradients to the same value matrix, so the source value must cover a larger union of content directions.
Thus cross-layer KV sharing amplifies the natural K/V specialization: keys become more compact retrieval indices, while values occupy richer content spaces, as observed in Fig.~\ref{fig:kv_erank}. In Appendix~\ref{app:theoretical_analysis}, we provide a detailed theoretical analysis that further supports the mechanism above under explicit assumptions.

\subsection{Connectivity Structure Determines Regularization Effectiveness}
\label{sec:prior_analysis}

Our analysis suggests that cross-layer gradients can act as a structural regularizer for KV geometry, but the regularization effect is determined by the topology of the sharing graph.
We now revisit existing KV-sharing designs and show that they expose complementary limitations.

\textbf{Adjacent-layer sharing: insufficient query-space separation.}
In CLA~\citep{brandon2024reducing} and MLKV~\citep{zuhri2025mlkv}, consecutive layers are grouped (typically groups of 2--3), and all layers within a group reuse the KV computed by the first layer.
For a source layer $s$ reused by its adjacent layer $s{+}1$, the source key receives
\begin{equation}
    \nabla_{\mathbf{K}_s}^{\mathrm{adj}}
    =
    \tfrac{1}{\sqrt{d}}\,\mathbf{G}_s^\top \mathbf{Q}_s
    +
    \tfrac{1}{\sqrt{d}}\,\mathbf{G}_{s+1}^\top \mathbf{Q}_{s+1}.
    \label{eq:grad_adjacent}
\end{equation}
This has the same form as Eq.~\ref{eq:grad_share}, but the two query spaces are nearly adjacent in depth.

\begin{wrapfigure}{r}{0.5\textwidth}
    \vspace{-12pt}
    \centering
    \begin{subfigure}[b]{0.24\textwidth}
        \centering
        \includegraphics[width=\textwidth]{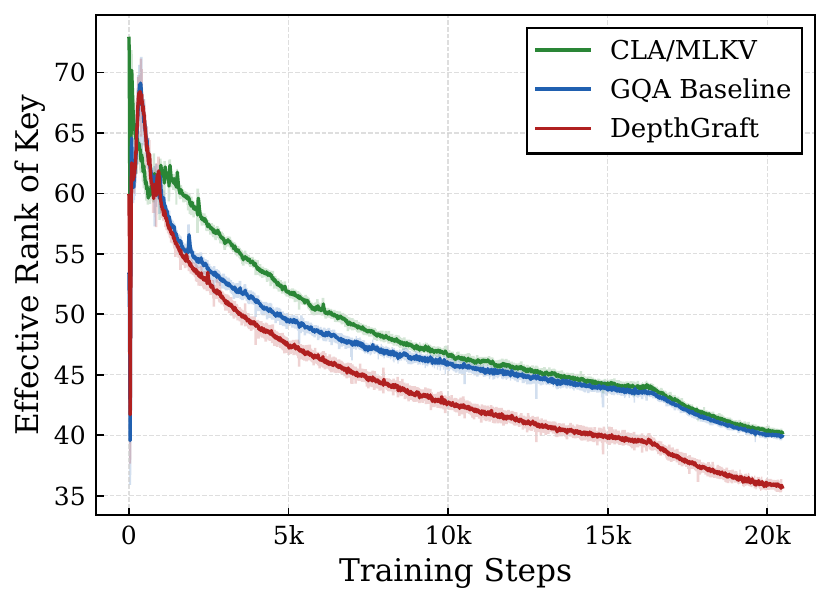}
        \caption{}
    \end{subfigure}
    \hfill
    \begin{subfigure}[b]{0.24\textwidth}
        \centering
        \includegraphics[width=\textwidth]{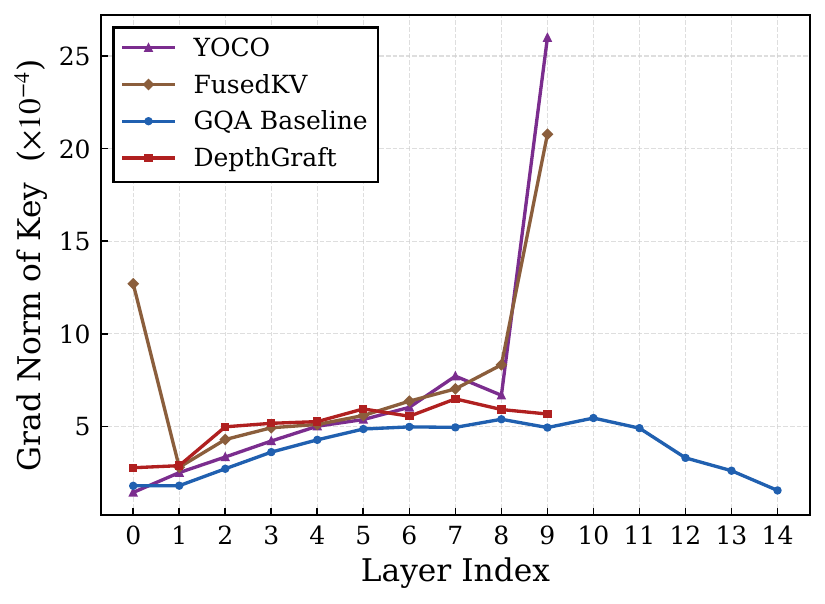}
        \caption{}
    \end{subfigure}
    \caption{(a) Adjacent-layer sharing (CLA/MLKV) does not reduce K erank below the baseline. (b) Global-anchor sharing concentrates gradient mass on anchor layers.}
    \label{fig:prior_analysis}
    \vspace{-10pt}
\end{wrapfigure}
Because neighboring hidden states are highly correlated, $\mathbf{Q}_s$ and $\mathbf{Q}_{s+1}$ provide largely overlapping retrieval signals.
The extra gradient path therefore adds little non-local pressure: it neither introduces enough query-space mismatch to strongly compress K toward shared discriminative directions nor provides sufficiently distinct attention patterns to expand V.
Fig.~\ref{fig:prior_analysis}(a) is consistent with this diagnosis: adjacent sharing does not produce the pronounced K-rank concentration predicted for useful cross-layer regularization.
In this regime, the representational constraint can dominate the weak regularization benefit: multiple layers attend to identical memory, potentially reducing the model's effective depth.

\textbf{Global-anchor sharing: gradient congestion and uneven regularization.}
In YOCO and FusedKV, a small number of early layers (one or two) serve as global anchors for all deep layers.
For an anchor layer $a$ shared by $M$ deep layers, the gradient accumulates contributions from every reuse layer:
\begin{equation}
    \nabla_{\mathbf{K}_a}^{\mathrm{global}} = \tfrac{1}{\sqrt{d}}\,\mathbf{G}_a^\top \mathbf{Q}_a \;+\; \sum_{j=1}^{M} \tfrac{1}{\sqrt{d}}\,\mathrm{diag}(\boldsymbol{\alpha}_j)\, \mathbf{G}_{\ell_j}^\top \mathbf{Q}_{\ell_j},
    \label{eq:grad_global}
\end{equation}
where $\alpha_j$ are per-layer coefficients (all ones for YOCO; learnable for FusedKV).
Compared with the two-way interaction in Eq.~\ref{eq:grad_share}, the anchor must satisfy $M$ heterogeneous query spaces at once.
This creates a high-fan-in multi-objective pressure on the same KV tensor: some cross-layer interference is useful, but concentrating all deep-layer objectives on one or two anchors can make the source representation an overloaded compromise.
Fig.~\ref{fig:prior_analysis}(b) shows this concentrated load directly: YOCO and FusedKV place most KV-gradient mass on their anchor layers.

Global-anchor sharing also makes the regularization structurally uneven.
Only the anchor layers receive direct cross-layer KV gradients from deep layers; the remaining shallow and middle layers may still receive gradients through the residual stream, but they do not participate in the direct KV-sharing path analyzed in Sec.~\ref{sec:analysis}.
Fig.~\ref{fig:v_erank_layer} in the appendix illustrates this effect for FusedKV: the anchor layer exhibits elevated V effective rank, whereas nearby non-anchor layers show no clear improvement over the baseline.

\subsection{\method{}: Hierarchical Cross-Layer KV Reconstruction}
\label{sec:depthgraft}

By viewing KV-sharing designs through the lens of cross-layer gradients as structural regularization, the analysis above motivates the following design principles:

\textbf{1. Sufficient cross-layer distance.} Source and reuse layers must be separated by enough depth that their query spaces meaningfully diverge, enabling the gradient interference that drives K/V specialization. Adjacent-layer sharing fails this condition.

\textbf{2. Distributed gradient flow with full-layer participation.} Each source layer should receive gradients from only a small number of reuse layers to avoid gradient congestion, and most layers should participate in the cross-layer connectivity graph so that the regularization benefit is not confined to a few anchors.

We instantiate these requirements under an efficiency constraint: all source KV tensors should be produced before the reconstructed layers, preserving prefill early completion.
Based on this, we propose \method{}, which introduces a \textbf{hierarchical stage layout} that partitions the $L$-layer Transformer into shallow, middle, and deep stages, each containing roughly $L/3$ layers.
The shallow and middle stages compute KV normally, while the deep stage reconstructs its KV from earlier stages.
Rather than using global anchors, \method{} adopts \textbf{stride-aligned source assignment}, treating deep-layer memory as a locally reconstructed trajectory over depth.
For each deep layer $\ell$, \method{} selects one shallow source and one middle source at the same relative depth position:
\begin{equation}
    s(\ell)=\ell-\tfrac{2L}{3}, \qquad
    m(\ell)=\ell-\tfrac{L}{3}.
    \label{eq:source_map}
\end{equation}

This gives each deep layer two memory bases at different abstraction levels: shallow memory preserves lower-level lexical and token-local patterns, while middle-layer memory carries more abstract semantic states~\cite{aljaafari2024mechanics,zheng2024attention,rocchetti2025instruction}.
It then reconstructs KV from the corresponding shallow--middle source pair with learnable per-dimension coefficients:
\begin{equation}
    \mathbf{K}_\ell = \mathbf{a}_{\ell,1} \odot \mathbf{K}_{s(\ell)} + \mathbf{a}_{\ell,2} \odot \mathbf{K}_{m(\ell)}, \qquad
    \mathbf{V}_\ell = \mathbf{b}_{\ell,1} \odot \mathbf{V}_{s(\ell)} + \mathbf{b}_{\ell,2} \odot \mathbf{V}_{m(\ell)},
    \label{eq:depthgraft}
\end{equation}
where $\mathbf{a}_{\ell,i}, \mathbf{b}_{\ell,i}$ are channel-wise coefficients for each KV group, providing higher flexibility and expressiveness than single-source reuse.
For keys, the coefficients are tied across RoPE rotation pairs, so rotary position encoding and reconstruction commute, which preserves the standard KV-cache storage logic.
The stride-aligned correspondence lets each deep layer receive shallow and middle memory from the same relative depth position, preserving the ordering of representation evolution across depth. Fig.~\ref{fig:depthgraft_workflow} in the appendix provides a workflow illustration of \method{}.

This design naturally yields the following advantages.

\emph{(1) Inference efficiency.}
Because the deep stage does not materialize independent KV, \method{} reduces the KV cache and prefill cost by approximately $1/3$.
Appendix~\ref{app:kv_cache_prefill} describes the cache-saving and prefill-early-completion mechanism.

\emph{(2) Systems friendliness.}
The stride-aligned source mapping is simple to implement and naturally compatible with pipeline parallelism (PP) and virtual pipeline parallelism (VPP), avoiding additional cross-stage KV communication under an appropriate placement.
Appendix~\ref{app:parallel_training} provides the corresponding PP/VPP details.

\emph{(3) Regularized KV geometry.}
The design satisfies the two principles above: non-local source--target pairs provide query-space separation, and stride alignment distributes cross-layer gradients across source layers.
As shown in Fig.~\ref{fig:kv_erank} and Fig.~\ref{fig:prior_analysis}(b), \method{} improves the geometry of KV representations while maintaining a healthy distributed gradient flow.
The effective-rank advantage in Fig.~\ref{fig:kv_erank} is observed across nearly all layers, rather than being confined to either source or reuse layers.

\section{Experiments}
\label{sec:experiments}

We evaluate \method{} on both dense and sparse MoE Transformers, covering multiple model scales and attention variants.
Our experiments are organized around three questions:
(1) whether the empirical results support the analysis in Sec.~\ref{sec:analysis}, i.e., whether \method{} can use cross-layer gradients as a structural regularizer to improve downstream performance while preserving the KV-cache and prefill-efficiency benefits of KV sharing;
(2) whether this effect scales to larger models and remains effective across different attention variants;
and (3) whether the cross-layer connectivity and stage-layout choices in Sec.~\ref{sec:depthgraft} are necessary, and how alternative sharing variants behave under the same training and evaluation protocol.

\subsection{Experimental Setup}
\label{sec:exp_setup}

\textbf{Training Settings.} We conduct experiments on both dense and sparse MoE language models.
For MoE models, we evaluate three scales of DeepSeekMoE-style Transformers~\cite{dai2024deepseekmoe} with SwiGLU feed-forward networks, ranging from 3B to 16B total parameters.
For dense models, we train models from 0.5B to 1.5B parameters.
The main comparisons use models trained from scratch with matched data and training recipes at each scale.
Training details are provided in Appendix~\ref{app:experimental_details}, and additional experiments are provided in Appendix~\ref{app:additional_results}.

\textbf{Evaluation.} We test few-shot performance on a broad downstream suite covering four capability groups:
knowledge and language understanding with MMLU~\cite{MMLU}, MMLU-Pro~\cite{wang2024mmlupro}, and LAMBADA~\cite{paperno2016lambada};
reasoning and mathematics with ARC-Challenge~\cite{ARCC}, SIQA~\cite{sap2019social}, and GSM8K~\cite{gsm8k};
code generation with HumanEval~\cite{humaneval}, MBPP~\cite{austin2021program}, and OpenCompass LCBench~\cite{2023opencompass};
and Chinese understanding with C-Eval~\cite{ceval}, CMMLU~\cite{li2024cmmlu}, and C3~\cite{c3}.

\subsection{Main Results}
\label{sec:exp_main}

\begin{table*}[t]
\centering
\caption{Dense scaling results of baseline and \method{} on 0.5B--1.5B models. We report accuracy or pass rate on each benchmark and average performance over all benchmarks. The best result within each model size is shown in \textbf{bold}.}
\label{tab:dense_scaling_main}
\resizebox{\textwidth}{!}{%
\begin{tabular}{llccccccccccccc}
\toprule
\multirow{2}{*}{\textbf{Model}} & \multirow{2}{*}{\textbf{Method}}
& \multicolumn{3}{c}{\textbf{Knowledge \& Understanding}}
& \multicolumn{3}{c}{\textbf{Reasoning \& Mathematics}}
& \multicolumn{3}{c}{\textbf{Code}}
& \multicolumn{3}{c}{\textbf{Chinese}}
& \multirow{2}{*}{\textbf{Avg.}} \\
\cmidrule(lr){3-5} \cmidrule(lr){6-8} \cmidrule(lr){9-11} \cmidrule(lr){12-14}
& & \textbf{MMLU} & \textbf{M-Pro} & \textbf{LAMB.} & \textbf{ARC-C} & \textbf{SIQA} & \textbf{GSM8K} & \textbf{HEval} & \textbf{MBPP} & \textbf{LCBench} & \textbf{C-Eval} & \textbf{CMMLU} & \textbf{C3} & \\
\midrule
\multirow{2}{*}{\textbf{0.5B}}
& GQA & 29.37 & 3.43 & 43.46 & 25.09 & \textbf{36.08} & 1.36 & 4.27 & 3.60 & \textbf{1.22} & \textbf{30.81} & 29.07 & 26.74 & 19.54 \\
& \method{} & \textbf{30.07} & \textbf{4.86} & \textbf{43.66} & \textbf{28.18} & 34.19 & \textbf{1.97} & \textbf{6.71} & \textbf{5.20} & 0.52 & 29.99 & \textbf{30.35} & \textbf{27.73} & \textbf{20.29} \\
\midrule
\multirow{2}{*}{\textbf{1B}}
& GQA & 30.35 & 7.86 & 52.58 & \textbf{31.27} & 35.26 & \textbf{6.67} & 9.15 & 9.40 & 2.09 & \textbf{31.45} & 30.23 & 31.67 & 23.17 \\
& \method{} & \textbf{32.70} & \textbf{8.57} & \textbf{54.17} & 27.84 & \textbf{38.08} & 6.07 & \textbf{9.76} & \textbf{12.60} & \textbf{2.61} & 31.43 & \textbf{30.52} & \textbf{33.92} & \textbf{24.02} \\
\midrule
\multirow{2}{*}{\textbf{1.5B}}
& GQA & 34.81 & 11.14 & 54.52 & 36.43 & 43.81 & 14.10 & 12.80 & 16.20 & 3.48 & \textbf{34.64} & 33.56 & 44.79 & 28.36 \\
& \method{} & \textbf{38.14} & \textbf{12.57} & \textbf{58.25} & \textbf{37.11} & \textbf{46.67} & \textbf{15.01} & \textbf{14.63} & \textbf{21.40} & \textbf{3.83} & 33.74 & \textbf{33.81} & \textbf{54.25} & \textbf{30.78} \\
\bottomrule
\end{tabular}%
}
\end{table*}

\begin{table*}[t]
\centering
\caption{MoE scaling results of baseline models and \method{} on 3B--16B models. We report accuracy or pass rate on each benchmark and average performance over all benchmarks. The best result within each model setting is shown in \textbf{bold}.}
\label{tab:moe_scaling_main}
\resizebox{\textwidth}{!}{%
\begin{tabular}{llccccccccccccc}
\toprule
\multirow{2}{*}{\textbf{Model}} & \multirow{2}{*}{\textbf{Method}}
& \multicolumn{3}{c}{\textbf{Knowledge \& Understanding}}
& \multicolumn{3}{c}{\textbf{Reasoning \& Mathematics}}
& \multicolumn{3}{c}{\textbf{Code}}
& \multicolumn{3}{c}{\textbf{Chinese}}
& \multirow{2}{*}{\textbf{Avg.}} \\
\cmidrule(lr){3-5} \cmidrule(lr){6-8} \cmidrule(lr){9-11} \cmidrule(lr){12-14}
& & \textbf{MMLU} & \textbf{M-Pro} & \textbf{LAMB.} & \textbf{ARC-C} & \textbf{SIQA} & \textbf{GSM8K} & \textbf{HEval} & \textbf{MBPP} & \textbf{LCBench} & \textbf{C-Eval} & \textbf{CMMLU} & \textbf{C3} & \\
\midrule
\multirow{2}{*}{\textbf{3B A0.5B}}
& GQA & 31.44 & \textbf{7.43} & 51.42 & 26.46 & 35.62 & 6.22 & 12.80 & \textbf{14.40} & 1.22 & 32.29 & 28.19 & 28.78 & 23.02 \\
& \method{} & \textbf{32.32} & 6.29 & \textbf{52.39} & \textbf{32.30} & \textbf{37.36} & \textbf{6.82} & \textbf{13.41} & 13.00 & \textbf{2.26} & \textbf{32.51} & \textbf{28.82} & \textbf{31.41} & \textbf{24.07} \\
\midrule
\multirow{2}{*}{\textbf{9B A1.5B}}
& GQA & 44.18 & \textbf{20.43} & 62.36 & 48.11 & 47.54 & 26.99 & \textbf{23.17} & \textbf{32.20} & 7.65 & 40.00 & 36.69 & 55.12 & 37.04 \\
& \method{} & \textbf{46.53} & 19.71 & \textbf{63.48} & \textbf{52.92} & \textbf{53.58} & \textbf{28.81} & 22.56 & 29.60 & \textbf{8.17} & \textbf{42.65} & \textbf{41.26} & \textbf{62.63} & \textbf{39.33} \\
\midrule
\multirow{2}{*}{\textbf{9B A1.5B}}
& MLA & 43.30 & 18.29 & \textbf{63.48} & 46.74 & 47.75 & 27.52 & 22.56 & \textbf{29.60} & 4.00 & 37.85 & 35.02 & 55.95 & 36.01 \\
& \method{} & \textbf{47.93} & \textbf{18.86} & 63.33 & \textbf{51.89} & \textbf{49.54} & \textbf{31.54} & \textbf{26.83} & 27.40 & \textbf{8.00} & \textbf{45.88} & \textbf{42.52} & \textbf{62.52} & \textbf{39.69} \\
\midrule
\multirow{2}{*}{\textbf{16B A3B}}
& GQA & 51.72 & 23.86 & 66.43 & 62.54 & 55.89 & 39.50 & 22.56 & 32.40 & 7.22 & 47.60 & 40.68 & 70.74 & 43.43 \\
& \method{} & \textbf{54.21} & \textbf{30.86} & \textbf{66.78} & \textbf{63.57} & \textbf{59.26} & \textbf{45.34} & \textbf{28.05} & \textbf{32.60} & \textbf{9.39} & \textbf{50.00} & \textbf{47.29} & \textbf{77.15} & \textbf{47.04} \\
\bottomrule
\end{tabular}%
}
\vspace{-10pt}
\end{table*}

\begin{table*}[t]
\centering
\caption{Long-context inference efficiency of baseline models and \method{}. We report time-to-first-token (TTFT) and time-per-output-token (TPOT) from 8K to 128K context lengths.}
\label{tab:efficiency_main}
\resizebox{0.8\textwidth}{!}{%
\begin{tabular}{llcccccccccc}
\toprule
\multirow{2}{*}{\textbf{Model}} & \multirow{2}{*}{\textbf{Method}}
& \multicolumn{5}{c}{\textbf{TTFT (ms)}}
& \multicolumn{5}{c}{\textbf{TPOT (ms)}} \\
\cmidrule(lr){3-7} \cmidrule(lr){8-12}
& & \textbf{8K} & \textbf{16K} & \textbf{32K} & \textbf{64K} & \textbf{128K}
& \textbf{8K} & \textbf{16K} & \textbf{32K} & \textbf{64K} & \textbf{128K} \\
\midrule
\multirow{2}{*}{\textbf{9B MLA}}
& Baseline & 52.22 & 89.05 & 161.75 & 314.09 & 611.19 & 3.39 & 3.54 & 3.84 & 4.09 & 4.59 \\
& \method{} & 43.82 & 69.89 & 122.77 & 233.31 & 447.51 & 3.42 & 3.57 & 3.87 & 4.13 & 4.62 \\
\midrule
\multirow{2}{*}{\textbf{16B GQA}}
& Baseline & 102.26 & 198.39 & 422.94 & 1000.15 & 2655.90 & 5.07 & 5.25 & 5.60 & 6.35 & 7.79 \\
& \method{} & 84.44 & 151.13 & 313.31 & 726.11 & 1891.80 & 5.12 & 5.30 & 5.64 & 6.40 & 7.84 \\
\bottomrule
\end{tabular}%
}
\vspace{-10pt}
\end{table*}

\textbf{Dense models.}
We first evaluate \method{} on dense Transformers at 0.5B, 1B, and 1.5B parameters (Table~\ref{tab:dense_scaling_main}).
\method{} improves the average score at all three scales, by +0.75, +0.85, and +2.42 points, respectively.
The gains are not confined to a single benchmark family.
At 1.5B, for example, \method{} improves knowledge understanding (MMLU +3.33, MMLU-Pro +1.43), reasoning and mathematics (SIQA +2.86, GSM8K +0.91), code generation (MBPP +5.20), and Chinese comprehension (C3 +9.46).
This consistent improvement on dense models indicates that constraining only the KV memory pathway does not reduce model capacity; instead, it provides a useful inductive bias for deep-layer memory representations.

\textbf{MoE models.}
We next scale \method{} to MoE architectures from 3B to 16B total parameters (Table~\ref{tab:moe_scaling_main}).
Across GQA-based MoE models, \method{} consistently outperforms the corresponding baseline, improving the average score by +1.05, +2.29, and +3.61 points at 3B, 9B, and 16B, respectively.
The 16B model shows especially strong gains, including MMLU-Pro (+7.00), CMMLU (+6.61), C3 (+6.41), and GSM8K (+5.84).
Furthermore, the consistent gains on both 9B GQA (+2.29 Avg.) and 9B MLA (+3.68 Avg.) further show that \method{} is not tied to a particular attention implementation.

Taken together, the main results support the mechanism developed in Sec.~\ref{sec:analysis}: cross-layer KV reconstruction improves performance not merely by saving cache, but by reshaping the geometry of KV representations through direct cross-layer gradient regularization.
The fact that the gains appear across dense models, sparse MoE models, different attention variants, and a wide range of scales suggests that the effect is a stable architectural bias rather than a scale-specific artifact.
During training, we also observe that the cross-layer gradient regularization introduced by \method{} can weaken attention-sink behavior, further supporting the view that \method{} improves the organization of the memory pathway.
Detailed visualization and analysis are provided in Appendix~\ref{app:attention_sink}.

\textbf{Efficiency Analysis.}
As discussed in Sec.~\ref{sec:depthgraft}, \method{} does not materialize independent KV tensors in the deep stage, yielding a fixed KV-cache reduction of roughly one third.
Table~\ref{tab:efficiency_main} further reports end-to-end TTFT and TPOT on 9B MLA and 16B GQA models with a fixed 2K generation length.
\method{} substantially reduces TTFT, and the gain becomes larger at longer contexts where prefill is increasingly compute-bound.
At 128K context length, TTFT is reduced by 26.8\% on 9B MLA and 28.8\% on 16B GQA.
In contrast, TPOT is nearly unchanged: the overhead remains below 1\% across all tested context lengths.
This suggests that \method{} preserves decoding efficiency while substantially reducing prefill latency, and the gap to the theoretical speedup can be further reduced with specialized kernel optimizations.

\subsection{Ablation Studies}
\label{sec:exp_ablation}

We conduct ablations on the 16B A3B MoE model to examine the roles of cross-layer gradient flow, connectivity pattern, and stage layout in \method{}.
We further compare \method{} with existing KV-sharing methods under the same training and evaluation protocol.

\textbf{Cross-layer gradients are the key source of the gain.}
Variants (2)--(4) all use the same three-stage layout, allowing us to isolate the effect of the cross-layer gradient path.
\emph{Shallow-Copy} directly reuses the stride-aligned shallow source from $\text{S}_1$, while \emph{Shallow-Copy (detach)} stops gradients through the reused KV.
The detached variant does not degrade relative to the GQA baseline (42.86 vs. 42.38 Avg.), indicating redundancy in deep-layer KV.
Once the backward path is enabled, however, the average score increases markedly to 45.30.
This controlled comparison supports our central motivation: KV sharing is useful not only as a forward memory reuse mechanism, but also because cross-layer gradients regularize earlier KV representations through deep-layer training signals.

\textbf{Hierarchical reconstruction is more effective than single-source copy.}
Variants (3)--(5) compare single-source copy with hierarchical reconstruction under active cross-layer gradients.
\emph{Shallow-Copy} in Variant (3) only connects $\text{S}_1$ to $\text{S}_3$, leaving the middle stage outside the sharing graph; the 2-stage variant (5) uses full one-to-one copy, but each reuse layer still receives only a single memory source.
\method{} instead reconstructs each deep-layer KV from a stride-aligned $\text{S}_1$--$\text{S}_2$ source pair, improving the average score over both single-source alternatives (47.56 vs. 45.30 and 43.58).
The gains on representative benchmarks such as M-Pro, C3, C-Eval, and GSM8K suggest that the benefit comes from combining broad source participation with a more flexible hierarchical memory basis, rather than from cross-layer gradient flow alone.

\textbf{The three-stage layout gives the best overall trade-off.}
Variants (5)--(8) compare different stage partitions under full gradient flow.
The reconstructed layers take stride-aligned sources and obtain their KV by either direct copy or hierarchical reconstruction.
The 2-stage variant is too restrictive, while the 4-stage variants test whether using more source stages or more reuse stages is beneficial.
The results suggest that reconstruction sources are not simply ``the more the better'': reconstructing the last stage from three source stages underperforms the proposed two-source DepthGraft layout, and the 4-stage two-source variant also remains below the three-stage design.
Thus, a three-stage partition provides a stronger operating point, balancing two complementary source abstraction levels, distributed gradient flow, and the intended final-stage cache-saving region.

\textbf{\method{} outperforms prior KV-sharing topologies.}
Table~\ref{tab:kv_sharing_comparison} compares \method{} with existing KV-sharing designs under the same 16B A3B MoE setting.
The results further support the topology analysis in Sec.~\ref{sec:prior_analysis}.
Adjacent sharing substantially underperforms the baseline, supporting the view that reusing nearby KV reduces the model's effective depth.
Global-anchor sharing performs better than adjacent sharing and can improve over the baseline, but shows limited gains because the gradient regularization is concentrated on a small set of anchors.
By distributing non-local source--reuse connections across layers, \method{} achieves the best overall performance.

\begin{table*}[t]
\centering
\caption{Ablation study on the 16B A3B MoE model. We decompose \method{} along two axes: cross-layer gradient connectivity and stage layout. Best results within each group are in \textbf{bold}.}
\label{tab:ablation_methods}
\small
\setlength{\tabcolsep}{3.5pt}
\resizebox{\textwidth}{!}{%
\begin{tabular}{lcccccccccc}
\toprule
\textbf{Method} & \textbf{Grad. Flow} & \textbf{Stage Layout}
& \textbf{MMLU} & \textbf{M-Pro} & \textbf{CMMLU} & \textbf{C3} & \textbf{C-Eval} & \textbf{GSM8K} & \textbf{HEval} & \textbf{Avg.} \\
\midrule
(1) GQA Baseline & -- & -- & 51.72 & 23.86 & 40.68 & 70.74 & 47.60 & 39.50 & 22.56 & 42.38 \\
\midrule
\multicolumn{11}{c}{\textit{Gradient Connectivity \& Hierarchical Reconstruction Variants}} \\
\midrule
(2) Shallow-Copy (detach) & \ding{55} & $\text{S}_1\to\text{S}_3$ & 49.72 & 24.14 & 42.08 & 70.96 & 44.76 & 41.55 & 26.83 & 42.86 \\
(3) Shallow-Copy & Partial & $\text{S}_1\to\text{S}_3$ & 52.53 & 29.43 & 46.72 & 70.96 & 45.50 & 43.29 & \textbf{28.66} & 45.30 \\
(4) \method{} & Full & $\text{S}_1\oplus\text{S}_2\to\text{S}_3$ & \textbf{54.21} & \textbf{30.86} & \textbf{47.29} & \textbf{77.15} & \textbf{50.00} & \textbf{45.34} & 28.05 & \textbf{47.56} \\
\midrule
\multicolumn{11}{c}{\textit{Stage Layout Variants}} \\
\midrule
(5) \method{} (2-stage) & Full & $\text{S}_1 \to \text{S}_2$ & 52.11 & 22.71 & \textbf{47.82} & 70.25 & 46.56 & 40.64 & 25.00 & 43.58 \\
(6) \method{} (4-stage, 3-src) & Full & $\text{S}_1\oplus\text{S}_2\oplus\text{S}_3\to\text{S}_4$ & 51.65 & 25.71 & 43.96 & 75.40 & 45.36 & \textbf{47.69} & \textbf{28.05} & 45.40 \\
(7) \method{} (4-stage, 2-src) & Full & $\text{S}_1\oplus\text{S}_2\to\text{S}_3,\text{S}_4$ & 53.26 & 27.14 & 47.43 & 72.05 & \textbf{50.48} & 43.14 & 23.78 & 45.33 \\
(8) \method{} & Full & $\text{S}_1\oplus\text{S}_2\to\text{S}_3$ & \textbf{54.21} & \textbf{30.86} & 47.29 & \textbf{77.15} & 50.00 & 45.34 & \textbf{28.05} & \textbf{47.56} \\
\bottomrule
\end{tabular}
}
\vspace{-5pt}
\end{table*}

\begin{table*}[t]
\centering 
\caption{Comparison with prior KV sharing methods on the 16B A3B MoE model. \emph{Type} indicates the sharing topology. Best results are in \textbf{bold}.}
\label{tab:kv_sharing_comparison}
\small
\setlength{\tabcolsep}{3.5pt}
\resizebox{0.8\textwidth}{!}{%
\begin{tabular}{llccccccccc}
\toprule
\textbf{Method} & \textbf{Type}
& \textbf{MMLU} & \textbf{M-Pro} & \textbf{CMMLU} & \textbf{C3} & \textbf{C-Eval} & \textbf{GSM8K} & \textbf{HEval} & \textbf{Avg.} \\
\midrule
GQA Baseline & -- & 51.72 & 23.86 & 40.68 & 70.74 & 47.60 & 39.50 & 22.56 & 42.38 \\
\midrule
CLA / MLKV & Adjacent & 48.81 & 20.14 & 40.64 & 58.25 & 40.50 & 37.91 & 20.12 & 38.05 \\
YOCO & Global-1 & 50.28 & 27.57 & 43.87 & 67.73 & 46.59 & 42.91 & 24.39 & 43.33 \\
FusedKV & Global-2 & 53.86 & 27.71 & 46.48 & 70.90 & \textbf{50.66} & 44.88 & 25.00 & 45.64 \\
\method{} & Hierarchical & \textbf{54.21} & \textbf{30.86} & \textbf{47.29} & \textbf{77.15} & 50.00 & \textbf{45.34} & \textbf{28.05} & \textbf{47.56} \\
\bottomrule
\end{tabular}
}
\vspace{-10pt}
\end{table*}

\subsection{Long-Context Evaluation}
\label{sec:long_context}

We further evaluate whether cross-layer KV reconstruction preserves
long-context retrieval. Both the baseline and \method{} 16B A3B models undergo
continued training at a 32K context length, with the RoPE base adjusted for
length extension. Table~\ref{tab:ruler} reports ten RULER~\cite{hsieh2024ruler}
tasks at context lengths from 4K to 32K, covering needle-in-a-haystack retrieval,
variable tracking, frequent-word extraction, and question answering.

\begin{table}[t]
\centering
\captionsetup{position=top,skip=7pt}
\caption{Long-context performance on RULER after context extension.
NIAH denotes needle-in-a-haystack
retrieval; VT, FWE, and QA denote variable tracking, frequent-word extraction,
and SQuAD-based question answering.}
\label{tab:ruler}
\resizebox{0.8\textwidth}{!}{%
\begin{tabular}{llrrrrrrrrrrr}
\toprule
\multirow{2}{*}{Length} & \multirow{2}{*}{Method}
& \multicolumn{7}{c}{NIAH} & \multirow{2}{*}{VT}
& \multirow{2}{*}{FWE} & \multirow{2}{*}{QA} & \multirow{2}{*}{Avg.} \\
\cmidrule(lr){3-9}
& & s1 & s2 & s3 & mk1 & mk3 & mv & mq & & & & \\
\midrule
\multirow{2}{*}{4K} & Baseline & 100 & 100 & 75.0 & 100 & 99.0 & 90.5 & 95.5 & 43.2 & 80.7 & 84.0 & 86.8 \\
& \method{} & 100 & 100 & 100 & 100 & 99.0 & 98.3 & 98.5 & 81.6 & 84.3 & 86.0 & \textbf{94.8} \\
\midrule
\multirow{2}{*}{8K} & Baseline & 100 & 100 & 70.0 & 97.0 & 96.0 & 58.3 & 95.3 & 44.0 & 65.0 & 66.0 & 79.2 \\
& \method{} & 100 & 100 & 100 & 94.0 & 100 & 98.0 & 94.5 & 80.0 & 78.7 & 61.0 & \textbf{90.6} \\
\midrule
\multirow{2}{*}{16K} & Baseline & 100 & 100 & 82.0 & 96.0 & 93.0 & 48.3 & 91.5 & 45.6 & 66.0 & 63.0 & 78.5 \\
& \method{} & 100 & 100 & 96.0 & 92.0 & 96.0 & 87.5 & 86.0 & 77.6 & 78.3 & 63.0 & \textbf{87.6} \\
\midrule
\multirow{2}{*}{32K} & Baseline & 100 & 100 & 99.0 & 95.0 & 61.0 & 66.3 & 87.8 & 37.8 & 64.0 & 51.0 & 76.2 \\
& \method{} & 100 & 100 & 100 & 94.0 & 76.0 & 73.5 & 76.5 & 76.6 & 74.7 & 53.0 & \textbf{82.4} \\
\bottomrule
\end{tabular}}
\end{table}

\method{} achieves a higher average at every evaluated context length,
including 94.8 vs. 86.8 at 4K and 82.4 vs. 76.2 at 32K.
The gains extend beyond simple retrieval: at 32K, variable tracking improves
from 37.8 to 76.6 and frequent-word extraction from 64.0 to 74.7.
It also matches or exceeds the baseline on most reported NIAH subtasks.
These results show that reconstructing deep-layer KV can retain and improve
long-context capabilities without storing independent KV at every layer.

\section{Conclusion and Future Work}
\label{sec:conclusion}

We presented \method{}, a hierarchical cross-layer KV reconstruction method that turns KV sharing from a cache-saving mechanism into a regularized memory pathway.
Our analysis shows that KV reuse creates direct cross-layer gradient paths that reshape KV representation geometry, and that the effectiveness of this regularization depends on the topology of the sharing graph.
Motivated by this view, \method{} reconstructs deep-layer KV from stride-aligned shallow--middle source pairs, preserving the efficiency benefits of KV sharing while distributing non-local gradient flow across source layers.
Experiments on dense and MoE models from 0.5B to 16B show consistent downstream gains, substantial reductions in KV cache and prefill cost, and negligible overhead during decoding.
Future work can develop a more rigorous theoretical understanding of how cross-layer KV sharing improves representation geometry, explore more effective sharing topologies, and further scale this direction to larger pretraining regimes.

\clearpage

{
\small

\bibliographystyle{plainnat}
\bibliography{reference}
}

\clearpage
\appendix

\section{Experimental Details}
\label{app:experimental_details}

\textbf{Model configurations.}
Tables~\ref{tab:app_dense_config} and~\ref{tab:app_moe_config} summarize the model configurations and training hyperparameters used in our dense and MoE experiments.
All models are decoder-only Transformers with SwiGLU feed-forward layers, RMSNorm, and RoPE positional embeddings.
The dense models use GQA throughout, while the MoE experiments cover both GQA and MLA variants. All MoE backbones follow a DeepSeekMoE-style layout~\cite{dai2024deepseekmoe}: the first Transformer layer is dense and the remaining layers are MoE layers.

\begin{table}[!htbp]
\centering
\small
\caption{Hyperparameters and model configurations for dense-model pretraining.}
\label{tab:app_dense_config}
\begin{tabular}{llccc}
\toprule
\textbf{Group} & \textbf{Parameter} & \textbf{0.5B} & \textbf{1B} & \textbf{1.5B} \\
\midrule
\multirow{8}{*}{Training}
& lr schedule & WSD & WSD & WSD \\
& lr & $8.0{\times}10^{-4}$ & $6.0{\times}10^{-4}$ & $6.0{\times}10^{-4}$ \\
& warmup steps & 2,000 & 2,000 & 2,000 \\
& decay ratio & 0.2 & 0.2 & 0.2 \\
& AdamW-$\beta$ & (0.9, 0.95) & (0.9, 0.95) & (0.9, 0.95) \\
& weight decay & 0.1 & 0.1 & 0.1 \\
& grad clip & 1.0 & 1.0 & 1.0 \\
& global batch size & 512 & 512 & -- \\
\midrule
\multirow{11}{*}{Model}
& hidden dim. & 1024 & 1024 & 1024 \\
& \# layers & 9 & 18 & 28 \\
& head dim. & 128 & 128 & 128 \\
& \# query heads & 8 & 8 & 8 \\
& \# KV heads & 4 & 4 & 4 \\
& context length & 8192 & 8192 & 8192 \\
& dense FFN size & 10944 & 10944 & 10944 \\
& \method{} reuse layers & 3/9 & 6/18 & 9/28 \\
& Normalization     & \multicolumn{3}{c}{RMSNorm ($\epsilon = 10^{-5}$)} \\
& Position Embedding   & \multicolumn{3}{c}{RoPE ($\theta = 50000$)} \\
\midrule
\multirow{2}{*}{Data}
& tokenizer & qwen2 & qwen2 & qwen2 \\
& \# tokens & -- & -- & -- \\
\bottomrule
\end{tabular}
\end{table}

\begin{table}[H]
\centering
\scriptsize
\caption{Hyperparameters and model configurations for MoE-model pretraining.}
\label{tab:app_moe_config}
\resizebox{0.85\textwidth}{!}{
\begin{tabular}{llcccc}
\toprule
\textbf{Group} & \textbf{Parameter} & \textbf{3B A0.5B} & \textbf{9B A1.5B} & \textbf{9B A1.5B} & \textbf{16B A3B} \\
 &  & \textbf{GQA} & \textbf{GQA} & \textbf{MLA} & \textbf{GQA} \\
\midrule
\multirow{8}{*}{Training}
& lr schedule & WSD & WSD & WSD & WS \\
& lr & $4.82{\times}10^{-4}$ & $4.82{\times}10^{-4}$ & $4.82{\times}10^{-4}$ & $4.2{\times}10^{-4}$ \\
& warmup steps & 2,000 & 2,000 & 2,000 & 750 \\
& decay ratio & 0.2 & 0.2 & 0.2 & -- \\
& AdamW-$\beta$ & (0.9, 0.95) & (0.9, 0.95) & (0.9, 0.95) & (0.9, 0.95) \\
& weight decay & 0.1 & 0.1 & 0.1 & 0.1 \\
& grad clip & 1.0 & 1.0 & 1.0 & 1.0 \\
& global batch size & 512 & -- & -- & -- \\
\midrule
\multirow{20}{*}{Model}
& hidden dim. & 768 & 2080 & 2080 & 2048 \\
& \# layers & 15 & 15 & 15 & 28 \\
& \# dense layers & 1 & 1 & 1 & 1 \\
& \# MoE layers & 14 & 14 & 14 & 27 \\
& head dim. & 128 & 128 & -- & 128 \\
& \# query heads & 8 & 16 & 16 & 16 \\
& \# KV heads & 4 & 8 & -- & 8 \\
& MLA q-lora rank & -- & -- & 1536 & -- \\
& MLA kv-lora rank & -- & -- & 384 & -- \\
& MLA qk head dim. & -- & -- & 80+48 & -- \\
& MLA v head dim. & -- & -- & 128 & -- \\
& context length & 8192 & 8192 & 8192 & 8192 \\
& dense FFN size & 10944 & 6144 & 6144 & 10944 \\
& \# routed experts & 144 & 144 & 144 & 64 \\
& topK route & 8 & 8 & 8 & 6 \\
& MoE FFN size & 544 & 704 & 704 & 1408 \\
& shared exp. size & 544 & 704 & 704 & 2816 \\
& \method{} reuse layers & 5/15 & 5/15 & 5/15 & 9/28 \\
& Normalization     & \multicolumn{4}{c}{RMSNorm ($\epsilon = 10^{-5}$)} \\
& Position Embedding   & \multicolumn{4}{c}{RoPE ($\theta = 50000$)} \\
\midrule
\multirow{2}{*}{Data}
& tokenizer & qwen2 & qwen2 & qwen2 & qwen2 \\
& \# tokens & -- & -- & -- & -- \\
\bottomrule
\end{tabular}
}
\end{table}

\textbf{Training hyperparameters.}
All comparisons at the same model scale use identical training recipes and token budgets, so that differences can be attributed to the KV-sharing design rather than optimization settings.
For the prior KV-sharing baselines in Table~\ref{tab:kv_sharing_comparison}, CLA/MLKV use three-layer sharing groups, while YOCO and FusedKV start reusing KV after the middle stage, i.e., after the first $2L/3$ layers, as illustrated in Fig.~\ref{fig:prior_analysis}(b).
This matches the reuse ratio of \method{} for YOCO and FusedKV, allowing us to compare performance under different connectivity patterns.
For our 28-layer models, we leave layer 0 unshared and apply the three-stage reconstruction to the remaining 27 layers, resulting in three 9-layer stages.
We use AdamW with $\beta_1{=}0.9$, $\beta_2{=}0.95$, weight decay 0.1, and gradient clipping at 1.0.
Models smaller than 16B use a warmup--stable--decay schedule~\cite{wen2024understanding}: the final 20\% of training tokens use exponential decay to reduce the learning rate to 0.1 of its peak value.
For the 16B MoE model, we use a warmup--stable schedule without LR decay and with batch-size ramp-up.
All models use the qwen2 tokenizer with a vocabulary size of 152,064 and are trained with a context length of 8192.
Our pretraining corpus is primarily proprietary and private, built from web text, code, math, and multilingual sources, with exact/fuzzy deduplication, rule-based filtering, and LLM-based quality filtering.

\textbf{Evaluation protocol.}
We evaluate all models with the same few-shot protocol under OpenCompass.
The benchmark suite covers knowledge and language understanding with MMLU (5-shot), MMLU-Pro (5-shot), and LAMBADA (0-shot); reasoning and mathematics with ARC-Challenge (25-shot), SIQA (5-shot), and GSM8K (4-shot); code generation with HumanEval (0-shot), MBPP (3-shot), and OpenCompass LCBench2023 (5-shot); and Chinese understanding with C-Eval (5-shot), CMMLU (5-shot), and C3 (3-shot).

\FloatBarrier
\section{Method Details and Additional Illustrations}
\label{app:method_details}

\begin{figure}[h]
    \centering
    \includegraphics[width=0.5\textwidth]{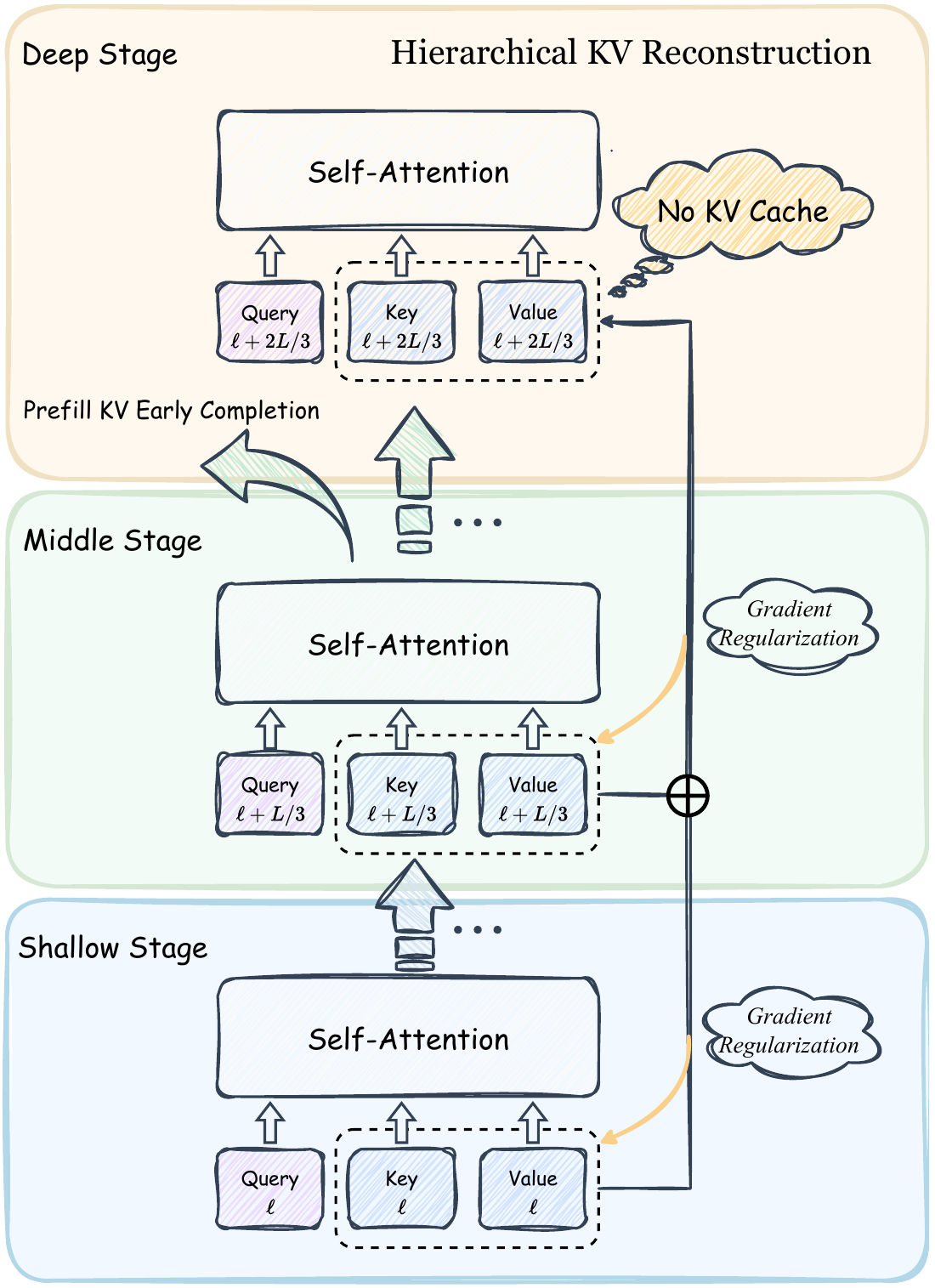}
    \caption{\method{} workflow. The shallow and middle stages compute and cache KV normally, while deep-stage KV is reconstructed from stride-aligned shallow--middle source pairs. The deep stage therefore does not materialize independent KV cache, while gradients from deep-layer attention flow back to the source KV projections through the reconstruction paths.}
    \label{fig:depthgraft_workflow}
\end{figure}

\textbf{RoPE-pair tying for K coefficients.}
Rotary position embeddings apply a block-diagonal rotation to each two-dimensional channel pair of the query and key vectors.
For the NeoX-style layout used in our implementation, dimensions $j$ and $j+d/2$ form one RoPE pair.
If a channel-wise coefficient assigns different values to the two dimensions in the same pair, then applying RoPE before and after reconstruction would no longer be equivalent.
Consistent with FusedKV~\citep{lin2025reconstructing}, we therefore tie the K reconstruction coefficients within each RoPE pair.
Concretely, for each source coefficient $\mathbf{a}_{\ell,i}$, we store a half-sized tensor $\mathbf{a}_{\ell,i}\in\mathbb{R}^{H_k\times d/2}$ and expand it to $\tilde{\mathbf{a}}_{\ell,i}\in\mathbb{R}^{H_k\times d}$ by
\begin{equation}
    \tilde{\mathbf{a}}_{\ell,i}[h,j]
    =
    \tilde{\mathbf{a}}_{\ell,i}[h,j+d/2]
    =
    \mathbf{a}_{\ell,i}[h,j],
    \qquad j=1,\ldots,d/2 .
\end{equation}
Let $\mathbf{R}_p$ denote the RoPE rotation matrix at position $p$.
Because each expanded coefficient is constant within every rotation pair, it commutes with the RoPE rotation:
\begin{equation}
    \mathbf{R}_p(\tilde{\mathbf{a}}_{\ell,i}\odot \mathbf{K})
    =
    \tilde{\mathbf{a}}_{\ell,i}\odot \mathbf{R}_p(\mathbf{K}).
\end{equation}
Therefore, for the reconstructed key,
\begin{equation}
\begin{aligned}
    \mathbf{R}_p\!\left(\tilde{\mathbf{a}}_{\ell,1}\odot \mathbf{K}_{s(\ell)}
    + \tilde{\mathbf{a}}_{\ell,2}\odot \mathbf{K}_{m(\ell)}\right)
    &=
    \tilde{\mathbf{a}}_{\ell,1}\odot \mathbf{R}_p(\mathbf{K}_{s(\ell)})
    + \tilde{\mathbf{a}}_{\ell,2}\odot \mathbf{R}_p(\mathbf{K}_{m(\ell)}).
\end{aligned}
\end{equation}
This means that key reconstruction can be performed either before or after RoPE without changing the result, preserving the standard KV-cache storage logic.
Value vectors do not use RoPE, so the V coefficients $\mathbf{b}_{\ell,i}$ are stored at full channel dimension and do not require pair tying.

\textbf{Coefficient initialization.}
All reconstruction coefficients $\mathbf{a}_{\ell,1},\mathbf{a}_{\ell,2},\mathbf{b}_{\ell,1},\mathbf{b}_{\ell,2}$ are initialized to $0.5$.
At the start of training, each deep layer therefore receives an equal-weight reconstruction from its shallow and middle sources.
This provides a stable interpolation between the two memory bases before the model learns layer- and channel-specific preferences.
As shown in Appendix~\ref{app:coefficient_dynamics}, the coefficients later deviate from this symmetric initialization and learn different source preferences for K and V.

\FloatBarrier
\section{Theoretical Analysis}
\label{app:theoretical_analysis}

\textbf{A conditional spectral view of cross-layer KV regularization.}
We provide a local spectral analysis to support the effective-rank diagnostic in Eq.~\ref{eq:erank}.
For a centered key or value activation matrix $\mathbf{M}$, denote its feature covariance by $\mathbf{C}_{\mathbf{M}}=\mathbf{M}^\top \mathbf{M}/N$.
The effective rank is high when the covariance spectrum is close to uniform, and low when activation energy is concentrated in a few feature directions.
Our goal is not to prove that cross-layer KV sharing always changes effective rank in a fixed direction.
Instead, we show that the observed K/V effective-rank trends follow under a natural spectral-alignment condition on the cross-layer update covariance.

\textbf{Lemma 1 (spectral injection and effective rank).}
Consider a local update of the activation matrix
\begin{equation}
    \mathbf{M}^+ = \mathbf{M}+\eta\boldsymbol{\Delta} ,
\end{equation}
where $\boldsymbol{\Delta}$ is the induced activation-space update and $\eta$ is the learning rate.
The covariance update admits the exact decomposition
\begin{equation}
\begin{aligned}
    \mathbf{C}_{\mathbf{M}^+}
    &=
    \frac{1}{N}(\mathbf{M}+\eta\boldsymbol{\Delta})^\top(\mathbf{M}+\eta\boldsymbol{\Delta})  \\
    &=
    \mathbf{C}_{\mathbf{M}}
    +
    \eta \mathbf{A}_{\mathbf{M},\boldsymbol{\Delta}}
    +
    \eta^2\boldsymbol{\Sigma}_{\boldsymbol{\Delta}},
    \label{eq:exact_cov_update}
\end{aligned}
\end{equation}
where
\begin{equation}
    \mathbf{A}_{\mathbf{M},\boldsymbol{\Delta}}
    =
    \frac{1}{N}(\mathbf{M}^\top\boldsymbol{\Delta}+\boldsymbol{\Delta}^\top \mathbf{M}),
    \qquad
    \boldsymbol{\Sigma}_{\boldsymbol{\Delta}}
    =
    \frac{1}{N}\boldsymbol{\Delta}^\top\boldsymbol{\Delta} .
\end{equation}
We analyze the directional trend over a short training window, where stochastic mini-batch fluctuations make the first-order alignment term $\mathbf{A}_{\mathbf{M},\boldsymbol{\Delta}}$ average to a small or low-rank correction that does not reverse the spectral ordering induced by $\boldsymbol{\Sigma}_{\boldsymbol{\Delta}}$.
Equivalently, for the purpose of identifying which feature directions receive persistent update energy, we approximate
\begin{equation}
    \mathbb{E}[\mathbf{C}_{\mathbf{M}^+}]
    \approx
    \mathbf{C}_{\mathbf{M}}+\eta^2\mathbb{E}[\boldsymbol{\Sigma}_{\boldsymbol{\Delta}}].
    \label{eq:second_order_cov_update}
\end{equation}
Let $\mathbf{C}_{\mathbf{M}}=\mathbf{U}\operatorname{diag}(\lambda_1,\ldots,\lambda_d)\mathbf{U}^\top$ with $\lambda_1\ge\cdots\ge\lambda_d\ge0$.
Assume that the update covariance is approximately aligned with the current representation geometry:
\begin{equation}
    \mathbf{U}^\top\mathbb{E}[\boldsymbol{\Sigma}_{\boldsymbol{\Delta}}]\mathbf{U}
    =
    \operatorname{diag}(\mu_1,\ldots,\mu_d)+\mathbf{R},
    \qquad
    \|\mathbf{R}\|_2 \ll \|\operatorname{diag}(\boldsymbol{\mu})\|_2 .
    \label{eq:approx_joint_diag}
\end{equation}
Ignoring the small off-diagonal residual $\mathbf{R}$, the eigenvalues evolve as
\begin{equation}
    \lambda_i^+
    =
    \lambda_i+\eta^2\mu_i
    =
    \lambda_i(1+\eta^2\rho_i),
    \qquad
    \rho_i=\mu_i/\lambda_i .
    \label{eq:eigen_relative_injection}
\end{equation}
After normalization, the effective-rank weights become
\begin{equation}
    p_i^+
    =
    \frac{\lambda_i^+}{\sum_j\lambda_j^+}
    =
    \frac{\lambda_i(1+\eta^2\rho_i)}
    {\sum_j \lambda_j(1+\eta^2\rho_j)} .
    \label{eq:normalized_spectral_update}
\end{equation}
If the relative injection $\rho_i$ is larger on the leading eigenspace than on the trailing eigenspace, then probability mass shifts toward already-dominant components, so $\mathbf{p}^+$ majorizes $\mathbf{p}$.
Since Shannon entropy is Schur-concave, $H(\mathbf{p}^+)<H(\mathbf{p})$ whenever the shift is nonzero, and therefore
\begin{equation}
    \operatorname{erank}(\mathbf{C}_{\mathbf{M}^+})<\operatorname{erank}(\mathbf{C}_{\mathbf{M}}).
\end{equation}
Conversely, if $\rho_i$ is larger on trailing or previously underused directions, the normalized spectrum becomes more uniform and
\begin{equation}
    \operatorname{erank}(\mathbf{C}_{\mathbf{M}^+})>\operatorname{erank}(\mathbf{C}_{\mathbf{M}}).
\end{equation}
This establishes that effective-rank changes are controlled by \emph{relative spectral injection}, not merely by the absolute magnitude of the update covariance.

\textbf{Cross-layer KV sharing.}
We now apply the lemma to a source layer $s$ whose KV activations are reused by a non-adjacent layer $\ell$.
Suppressing head indices and reconstruction coefficients for clarity, the source KV gradients take the form
\begin{equation}
    \nabla_{\mathbf{K}_s}^{\mathrm{share}}
    =
    \frac{1}{\sqrt d}
    \left(\mathbf{G}_s^\top \mathbf{Q}_s+\mathbf{G}_\ell^\top \mathbf{Q}_\ell\right),
    \qquad
    \nabla_{\mathbf{V}_s}^{\mathrm{share}}
    =
    \mathbf{P}_s^\top \mathbf{E}_s+\mathbf{P}_\ell^\top \mathbf{E}_\ell ,
    \label{eq:share_grad_prop}
\end{equation}
where
\begin{equation}
    \mathbf{G}_t=\frac{\partial\mathcal{L}}{\partial(\mathbf{Q}_t\mathbf{K}_s^\top/\sqrt d)},
    \qquad
    \mathbf{E}_t=\frac{\partial\mathcal{L}}{\partial(\mathbf{P}_t\mathbf{V}_s)},
    \qquad t\in\{s,\ell\}.
\end{equation}
Under a local linearization of the projection update, we assume that the induced activation update covariance follows the covariance of these activation gradients.
The key-side covariance contains local and cross-layer terms:
\begin{equation}
\begin{aligned}
    \boldsymbol{\Sigma}_K^{\mathrm{share}}
    &\propto
    \mathbb{E}\Big[
    \mathbf{Q}_s^\top \mathbf{G}_s\mathbf{G}_s^\top \mathbf{Q}_s
    +
    \mathbf{Q}_\ell^\top \mathbf{G}_\ell \mathbf{G}_\ell^\top \mathbf{Q}_\ell  \\
    &\hspace{4.0em}
    +
    \mathbf{Q}_s^\top \mathbf{G}_s\mathbf{G}_\ell^\top \mathbf{Q}_\ell
    +
    \mathbf{Q}_\ell^\top \mathbf{G}_\ell \mathbf{G}_s^\top \mathbf{Q}_s
    \Big].
    \label{eq:key_cov_share}
\end{aligned}
\end{equation}
The value-side covariance similarly expands as
\begin{equation}
\begin{aligned}
    \boldsymbol{\Sigma}_V^{\mathrm{share}}
    &\propto
    \mathbb{E}\Big[
    \mathbf{E}_s^\top \mathbf{P}_s\mathbf{P}_s^\top \mathbf{E}_s
    +
    \mathbf{E}_\ell^\top \mathbf{P}_\ell \mathbf{P}_\ell^\top \mathbf{E}_\ell  \\
    &\hspace{4.0em}
    +
    \mathbf{E}_s^\top \mathbf{P}_s\mathbf{P}_\ell^\top \mathbf{E}_\ell
    +
    \mathbf{E}_\ell^\top \mathbf{P}_\ell \mathbf{P}_s^\top \mathbf{E}_s
    \Big].
    \label{eq:value_cov_share}
\end{aligned}
\end{equation}

\textbf{Proposition 1 (asymmetric effective-rank pressure from cross-layer KV sharing).}
Suppose that (i) cross-layer key gradients share a low-dimensional task-relevant retrieval subspace $\mathcal{U}_K$ aligned with the leading eigenspace of $\mathbf{C}_{\mathbf{K}_s}$, and their cross-covariance is coherent inside $\mathcal{U}_K$ but weak outside it; and (ii) value readout gradients from the source and reuse layers contain complementary content directions with non-negligible projection onto trailing or underused eigendirections of $\mathbf{C}_{\mathbf{V}_s}$.
Then, under the spectral approximation in Eqs.~\ref{eq:second_order_cov_update}--\ref{eq:approx_joint_diag}, cross-layer KV sharing locally induces
\begin{equation}
    \operatorname{erank}(\mathbf{C}_{\mathbf{K}_s^+})
    <
    \operatorname{erank}(\mathbf{C}_{\mathbf{K}_s}),
    \qquad
    \operatorname{erank}(\mathbf{C}_{\mathbf{V}_s^+})
    >
    \operatorname{erank}(\mathbf{C}_{\mathbf{V}_s}).
    \label{eq:kv_erank_prop}
\end{equation}

\textit{Proof.}
For keys, Eq.~\ref{eq:key_cov_share} shows that cross-layer sharing adds covariance terms coupling $\mathbf{G}_s^\top \mathbf{Q}_s$ and $\mathbf{G}_\ell^\top \mathbf{Q}_\ell$.
By assumption, these terms reinforce one another inside $\mathcal{U}_K$, producing a positive spectral increment on directions shared by the two query spaces.
Outside $\mathcal{U}_K$, the cross-layer components are weakly correlated, so their contribution to trailing eigenvalues is small relative to the coherent leading injection.
Therefore the relative increments satisfy
\begin{equation}
    \frac{\mu_i^K}{\lambda_i^K}
    >
    \frac{\mu_j^K}{\lambda_j^K},
    \qquad
    i\in\mathcal{I}_{\mathrm{lead}}(\mathbf{K}_s),\;
    j\in\mathcal{I}_{\mathrm{trail}}(\mathbf{K}_s),
    \label{eq:key_relative_condition}
\end{equation}
where $\mu_i^K$ denotes the key update-covariance eigenvalue in the approximate eigenbasis of $\mathbf{C}_{\mathbf{K}_s}$.
By Lemma~1, the normalized key covariance spectrum becomes more concentrated, yielding
\begin{equation}
    \operatorname{erank}(\mathbf{C}_{\mathbf{K}_s^+})
    <
    \operatorname{erank}(\mathbf{C}_{\mathbf{K}_s}).
\end{equation}

For values, Eq.~\ref{eq:value_cov_share} shows that the reuse layer routes an additional readout gradient $\mathbf{P}_\ell^\top \mathbf{E}_\ell$ into the same source value matrix.
Unlike keys, values are read out linearly once the attention pattern is fixed; distinct attention patterns can therefore provide complementary content gradients rather than competing through the softmax logits.
By assumption, these additional content gradients have non-negligible projection onto underused value directions, so
\begin{equation}
    \frac{\mu_i^V}{\lambda_i^V}
    <
    \frac{\mu_j^V}{\lambda_j^V},
    \qquad
    i\in\mathcal{I}_{\mathrm{lead}}(\mathbf{V}_s),\;
    j\in\mathcal{I}_{\mathrm{trail}}(\mathbf{V}_s).
    \label{eq:value_relative_condition}
\end{equation}
Lemma~1 then implies that the normalized value covariance spectrum becomes more uniform:
\begin{equation}
    \operatorname{erank}(\mathbf{C}_{\mathbf{V}_s^+})
    >
    \operatorname{erank}(\mathbf{C}_{\mathbf{V}_s}).
\end{equation}
Combining the two parts proves Eq.~\ref{eq:kv_erank_prop}.
\hfill$\square$

\textbf{Interpretation and scope.}
The proposition is conditional: it does not claim that effective rank must change monotonically under arbitrary training dynamics.
It states that the K/V effective-rank trends follow when the induced update covariance is organized as suggested by the attention mechanism: keys receive coherent reinforcement in shared retrieval directions, while values receive complementary content gradients.
The topology of the sharing graph affects whether these assumptions are plausible.
Adjacent sharing may make $\mathbf{Q}_s\approx \mathbf{Q}_\ell$ and $\mathbf{P}_s\approx \mathbf{P}_\ell$, yielding mostly redundant gradients.
Global-anchor sharing can overload a few sources with many heterogeneous objectives.
Stride-aligned non-local sharing is more likely to satisfy the assumptions because it provides sufficient cross-layer distance while distributing the gradient load across source layers.

\section{Additional Experimental Results}
\label{app:additional_results}

\subsection{Retention of Gains after Supervised Fine-tuning}
\label{app:sft}

We apply supervised fine-tuning (SFT) to the baseline and \method{}
16B A3B pretrained models using the same data, learning-rate schedule, and
update counts, and evaluate both with the same downstream protocol.
Table~\ref{tab:sft} shows that the pretraining advantage persists after SFT:
\method{} achieves an average score of 52.00, compared with 48.52 for the
baseline. It improves on 11 benchmarks and matches the baseline on LCBench,
with notable gains on MBPP (+7.00), HumanEval (+5.49), ARC-C (+4.81), and
GSM8K (+2.96). The gains from KV reconstruction therefore carry through
supervised adaptation rather than being erased by post-training.

\begin{table}[!htbp]
\centering
\caption{Downstream performance after SFT on the 16B A3B models.
Both models use the same fine-tuning recipe.}
\label{tab:sft}
\resizebox{\textwidth}{!}{%
\begin{tabular}{lccccccccccccc}
\toprule
Method & MMLU & M-Pro & LAMB. & ARC-C & SIQA & GSM8K & HEval & MBPP & LCBench & C-Eval & CMMLU & C3 & Avg. \\
\midrule
Baseline + SFT & 53.72 & 35.00 & 60.77 & 67.01 & 59.83 & 52.08 & 37.80 & 35.80 & \textbf{14.96} & 49.46 & 43.47 & 72.33 & 48.52 \\
\method{} + SFT & \textbf{56.84} & \textbf{37.57} & \textbf{62.28} & \textbf{71.82} & \textbf{62.54} & \textbf{55.04} & \textbf{43.29} & \textbf{42.80} & \textbf{14.96} & \textbf{52.03} & \textbf{47.28} & \textbf{77.59} & \textbf{52.00} \\
\bottomrule
\end{tabular}}
\end{table}

\FloatBarrier
\subsection{Compatibility with Sliding-Window Attention}
\label{app:swa}

\method{} changes the source of deep-layer KV, while sliding-window
attention (SWA) restricts the positions that each query can attend to.
We test their combination on a 12-layer MoE model with 8B total and
approximately 1B activated parameters, using a 3:1 ratio of SWA to full-attention
layers and a window size of 512. KV reuse is restricted to layers of the same
attention type: SWA layers reuse SWA sources, and full-attention layers reuse
full-attention sources. This avoids mixing the two types, whose KV-gradient
norms differ in our experiments, while preserving prefill early completion
because all sources remain in the first two stages.

\begin{table}[!htbp]
\centering
\caption{Compatibility with SWA on an 8B MoE model.
The two variants share the same attention-mask pattern and training recipe.}
\label{tab:swa}
\resizebox{\textwidth}{!}{%
\begin{tabular}{lccccccccccccc}
\toprule
Method & MMLU & M-Pro & LAMB. & ARC-C & SIQA & GSM8K & HEval & MBPP & LCBench & C-Eval & CMMLU & C3 & Avg. \\
\midrule
SWA baseline & 37.05 & 11.71 & 60.26 & 21.62 & \textbf{43.67} & 16.38 & \textbf{21.34} & 22.20 & \textbf{3.13} & \textbf{34.34} & 34.32 & 31.45 & 28.12 \\
SWA + \method{} & \textbf{38.25} & \textbf{13.57} & \textbf{61.43} & \textbf{33.33} & 40.31 & \textbf{19.50} & 18.90 & \textbf{23.00} & 2.57 & 33.57 & \textbf{36.54} & \textbf{35.63} & \textbf{29.72} \\
\bottomrule
\end{tabular}}
\end{table}

As shown in Table~\ref{tab:swa}, \method{} improves the average score
from 28.12 to 29.72, with gains on ARC-C, GSM8K, and C3 among other tasks.
These results support combining cross-layer KV reconstruction with local
attention, extending its applicability beyond full-attention backbones.
An alternative extension is a causal encoder--decoder design inspired by
DeepSeek-V4.1-Flash~\citep{xu2026deepseek}, applying KV reconstruction only
to full-attention layers while allowing SWA layers to retain their own KV.
This gives local representations greater freedom to evolve across layers
and could preserve the benefits of prefill early completion through SWA
bounded replay over a short suffix of the prompt.

\FloatBarrier
\section{Additional Training Dynamics and Representation Analysis}
\label{app:training_dynamics}

\subsection{Effective Rank Dynamics: Per-Layer Analysis}
\label{app:effective_rank_dynamics}

Sec.~\ref{sec:prior_analysis} argues that global-anchor sharing, such as FusedKV, induces structurally uneven regularization: only the designated anchor layers participate in cross-layer gradient flow, while nearby non-anchor layers receive no direct gradient benefit from KV reuse.
We provide per-layer empirical evidence for this effect by tracking the value effective rank $\operatorname{erank}(\mathbf{V}_\ell)$, defined in Eq.~\ref{eq:erank}, at individual layers throughout training.

\begin{figure}[h]
    \centering
    \begin{subfigure}[b]{0.24\textwidth}
        \includegraphics[width=\textwidth]{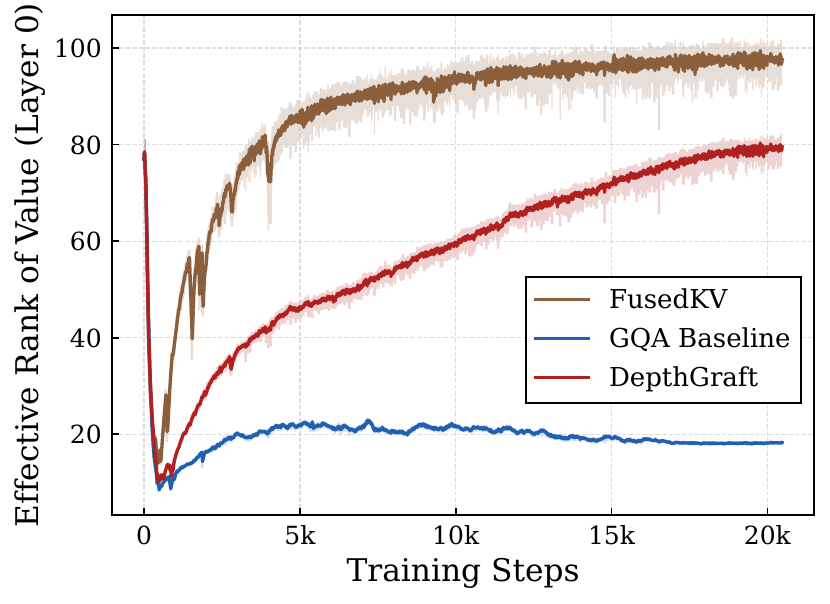}
        \caption{Layer 0}
    \end{subfigure}
    \hfill
    \begin{subfigure}[b]{0.24\textwidth}
        \includegraphics[width=\textwidth]{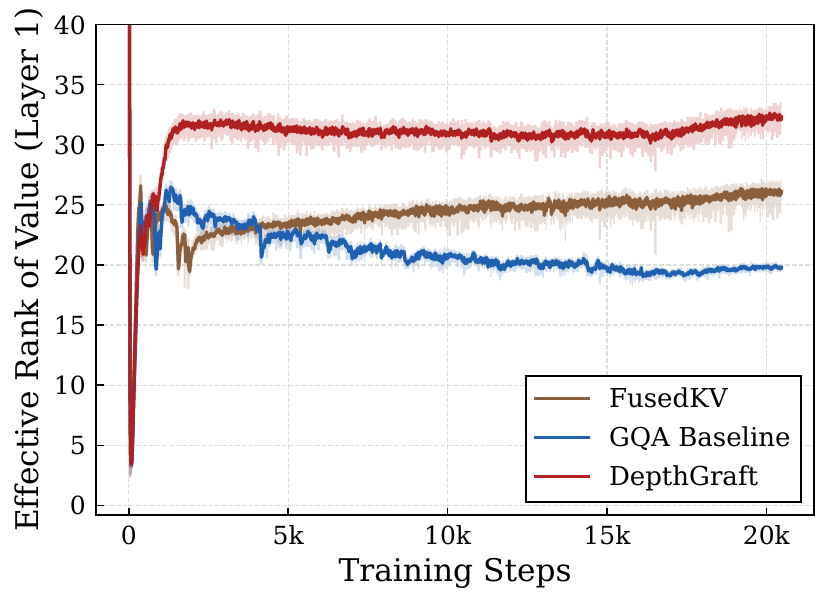}
        \caption{Layer 1}
    \end{subfigure}
    \hfill
    \begin{subfigure}[b]{0.24\textwidth}
        \includegraphics[width=\textwidth]{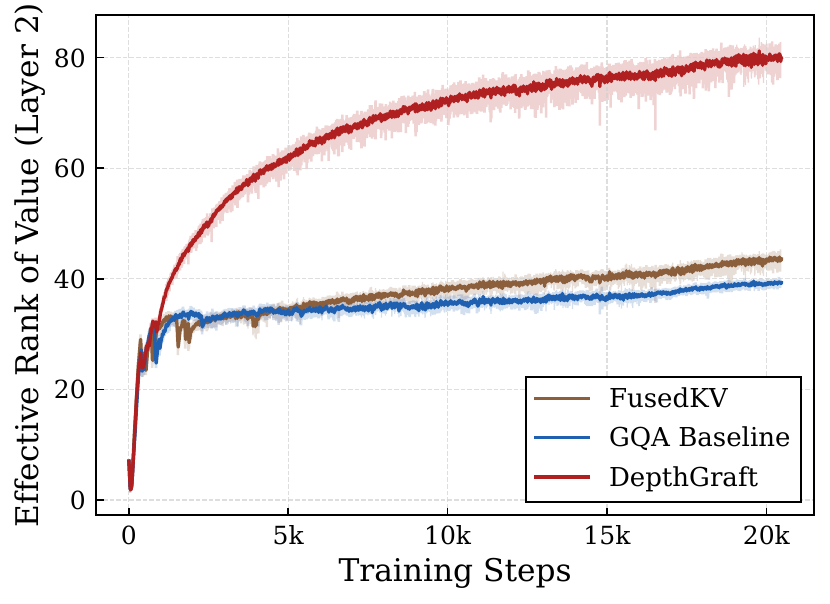}
        \caption{Layer 2}
    \end{subfigure}
    \hfill
    \begin{subfigure}[b]{0.24\textwidth}
        \includegraphics[width=\textwidth]{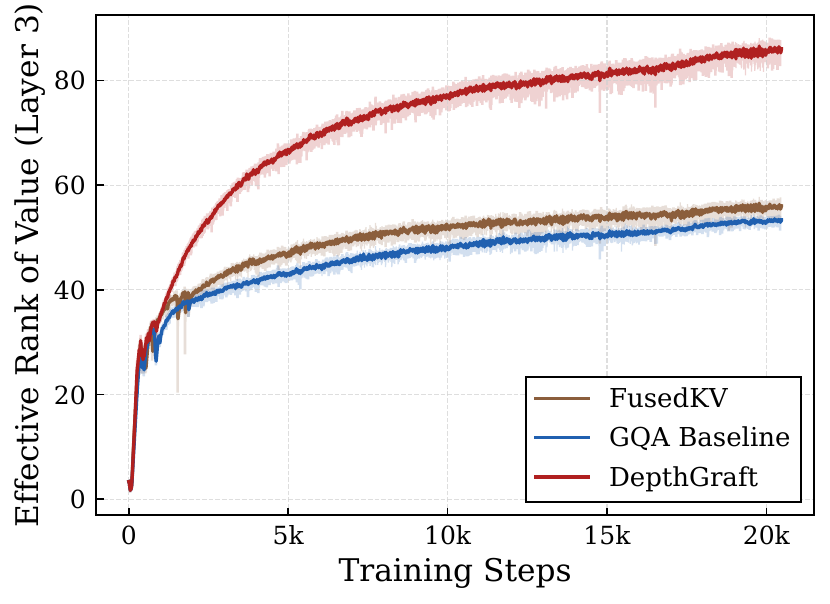}
        \caption{Layer 3}
    \end{subfigure}
    \caption{Per-layer V effective rank during training for the 9B GQA baseline, FusedKV, and \method{}. FusedKV improves the anchor layer but leaves nearby non-anchor layers close to the baseline, whereas \method{} yields consistent rank elevation across the shown source layers due to distributed stride-aligned connectivity.}
    \label{fig:v_erank_layer}
\end{figure}

Figure~\ref{fig:v_erank_layer} shows V effective-rank trajectories for layers 0--3 under the 9B GQA baseline, FusedKV, and \method{}.
For FusedKV, only layer~0, the bottom anchor, exhibits a clear increase above the baseline.
Layers 1--3 do not serve as anchors and therefore receive no direct cross-layer gradient through reused KV tensors; correspondingly, their V effective ranks show no clear improvement over the baseline.
This supports the topology analysis in Sec.~\ref{sec:prior_analysis}: global-anchor sharing confines the representation benefit to a small set of source layers.

In contrast, \method{} elevates V effective rank across the shown source layers.
Because its stride-aligned source assignment distributes reuse edges across the shallow and middle stages, the gradient-mediated regularization described in Sec.~\ref{sec:analysis} is spread over many source layers rather than concentrated on a few anchors.
The per-layer effective-rank measurements therefore provide additional evidence that distributed cross-layer connectivity is important for turning KV sharing into an effective representation regularizer.

\subsection{Attention Sink Analysis}
\label{app:attention_sink}

A common pathology in deep Transformers is the \emph{attention sink} phenomenon, where a disproportionate amount of attention mass is assigned to the first token regardless of its semantic relevance~\citep{xiao2023efficient}.
In our setting, this behavior provides an additional diagnostic of key geometry: if a layer develops extreme key directions, the softmax distribution can become dominated by a small number of sink positions.

\method{} mitigates this tendency through cross-layer gradient regularization.
Each shallow or middle source serves its own attention layer while also being reused by a stride-aligned deep layer, so its KV projections receive optimization signals from multiple query spaces.
This structure can regularize the source key geometry and make it less likely for a layer to develop sink-prone key directions that dominate the softmax.
This is consistent with the analysis in Sec.~\ref{sec:analysis}, where useful cross-layer KV sharing makes keys form a more compact retrieval geometry.

\begin{figure}[h]
\centering
\begin{subfigure}[t]{0.48\textwidth}
    \centering
    \includegraphics[width=\linewidth]{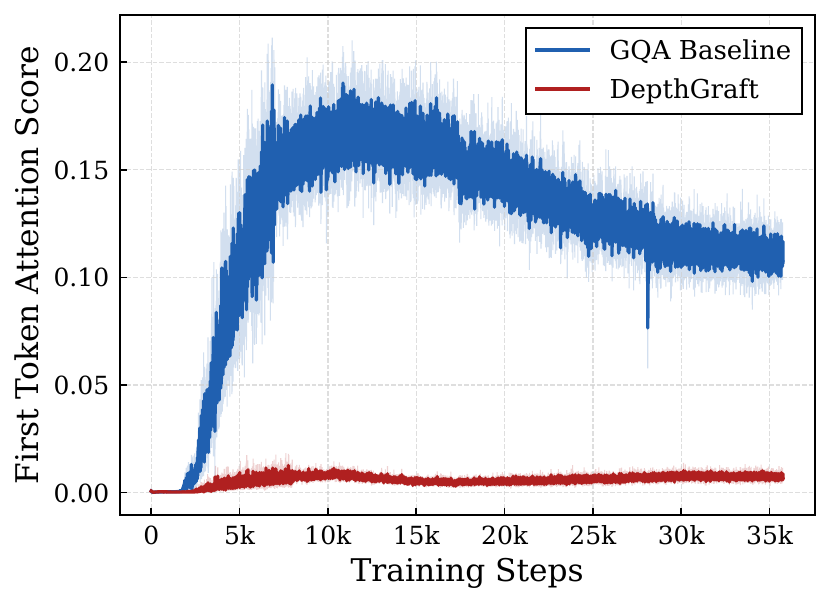}
    \caption{9B GQA model.}
    \label{fig:attn_sink_gqa}
\end{subfigure}
\hfill
\begin{subfigure}[t]{0.48\textwidth}
    \centering
    \includegraphics[width=\linewidth]{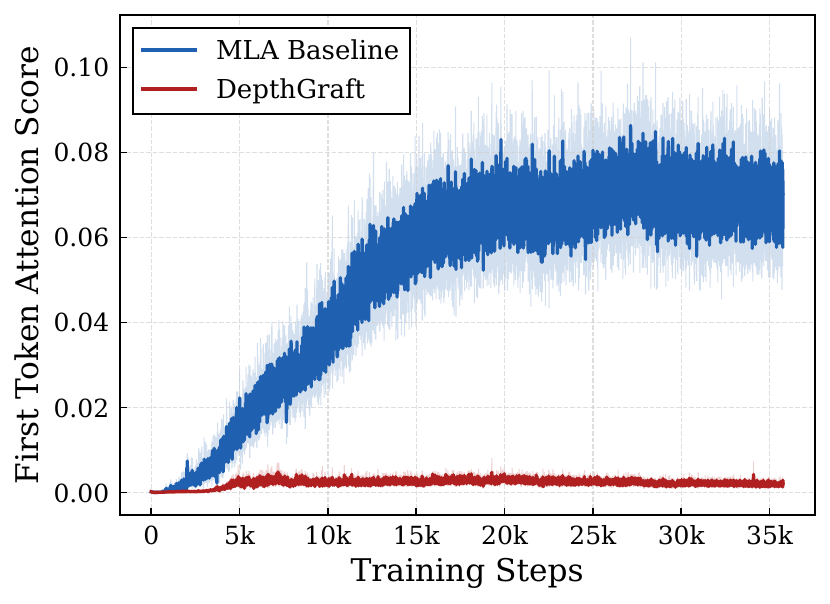}
    \caption{9B MLA model.}
    \label{fig:attn_sink_mla}
\end{subfigure}
\caption{Average attention score assigned to the first token over training steps. \method{} consistently reduces first-token attention concentration compared with the baseline on both GQA and MLA models.}
\label{fig:attn_sink}
\end{figure}

We verify this empirically by tracking the average attention score assigned to the first token throughout training.
Figure~\ref{fig:attn_sink} reports this metric for both 9B GQA and 9B MLA models.
In both cases, \method{} consistently lowers first-token attention concentration relative to the baseline, indicating reduced attention-sink behavior.

\subsection{Training Loss Curves}
\label{app:loss_curves}

\begin{figure}[h]
\centering
\begin{subfigure}[t]{0.32\textwidth}
    \centering
    \includegraphics[width=\linewidth]{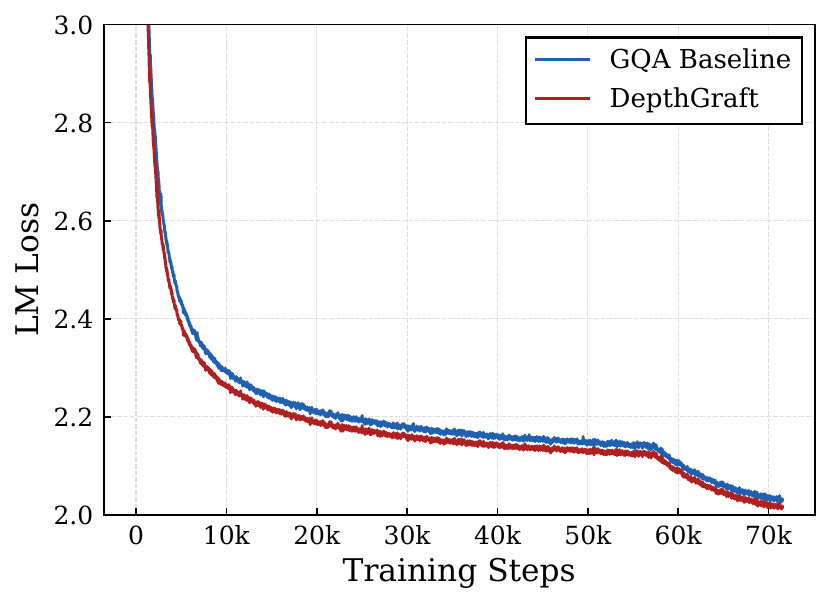}
    \caption{1.5B Dense.}
\end{subfigure}
\hfill
\begin{subfigure}[t]{0.32\textwidth}
    \centering
    \includegraphics[width=\linewidth]{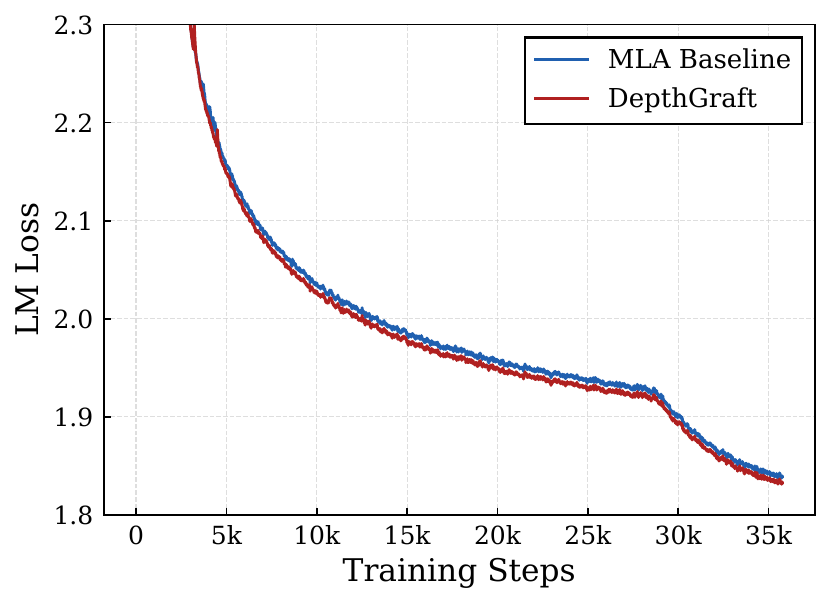}
    \caption{9B MoE MLA.}
\end{subfigure}
\hfill
\begin{subfigure}[t]{0.32\textwidth}
    \centering
    \includegraphics[width=\linewidth]{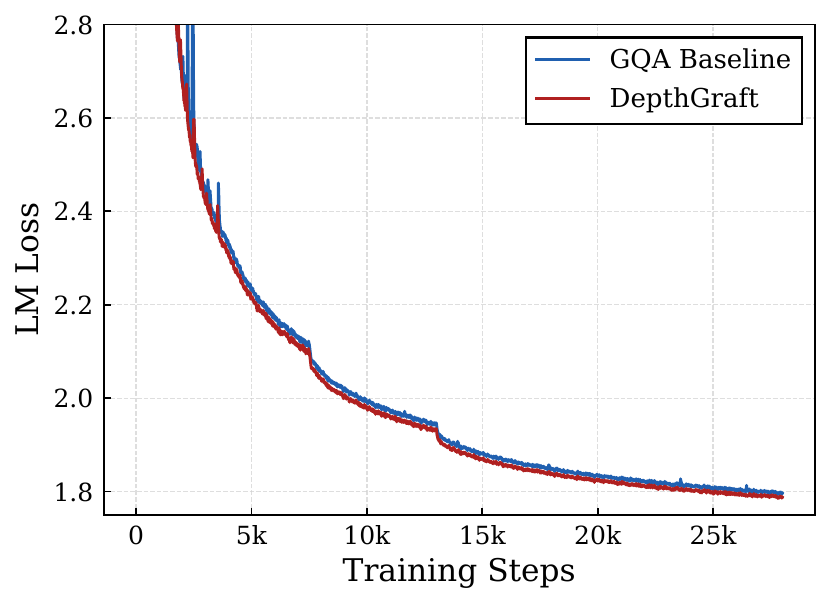}
    \caption{16B MoE GQA.}
\end{subfigure}
\caption{Training loss curves across representative model scales and attention variants. \method{} consistently improves training loss, but the gain is modest relative to its downstream benchmark improvements, supporting its interpretation as a KV-geometry regularizer.}
\label{fig:loss_curves}
\end{figure}

Figure~\ref{fig:loss_curves} compares the training loss curves of \method{} and the corresponding baselines across representative dense and MoE settings.
\method{} consistently achieves lower training loss across model scales and attention variants, indicating that the proposed cross-layer KV reconstruction does not introduce optimization instability.
The magnitude of the loss improvement, however, is moderate: the relative loss reduction is roughly $0.5\%$.

This behavior is consistent with the role of \method{} as a structural regularizer.
Unlike cross-layer designs whose primary effect is to substantially alter the residual computation pathway, \method{} constrains only the KV memory pathway and even removes independent deep-layer KV projections.
Consequently, its training-loss advantage is not as large as that reported for some broader cross-layer connectivity mechanisms, but it yields much larger downstream gains than the loss difference alone would suggest.
This supports our interpretation that \method{} improves the geometry and generalization of KV representations, rather than merely optimizing the pretraining loss more aggressively.

\subsection{Reconstruction Coefficient Dynamics}
\label{app:coefficient_dynamics}

We further inspect the learned channel-wise reconstruction coefficients.
For each reused layer, \method{} reconstructs deep-layer KV from a shallow source (src1) and a middle source (src2).
Table~\ref{tab:coeff_dynamics} reports the final coefficient norm normalized by its initialization norm; values larger than 1 indicate that the corresponding source receives increased weight during training.

\begin{table}[!htbp]
\centering
\small
\caption{Final reconstruction-coefficient norm relative to initialization. Src1 denotes the shallow source and src2 denotes the middle source. Values larger than 1 indicate increased reliance on the corresponding source after training.}
\label{tab:coeff_dynamics}
\begin{tabular}{ccccc}
\toprule
\textbf{Layer} & \textbf{K src1} & \textbf{K src2} & \textbf{V src1} & \textbf{V src2} \\
\midrule
10 & 0.6905 & 1.1955 & 2.1655 & 0.7594 \\
11 & 0.4738 & 1.1936 & 1.6323 & 1.4713 \\
12 & 0.5894 & 1.2603 & 2.4572 & 1.2290 \\
13 & 0.7178 & 1.0534 & 2.6295 & 0.8892 \\
14 & 0.7208 & 0.9557 & 1.8670 & 0.6671 \\
\bottomrule
\end{tabular}
\end{table}

The learned coefficients reveal a consistent K/V asymmetry.
For keys, the middle source is generally upweighted relative to the shallow source, suggesting that deep-layer retrieval benefits more from the more abstract middle-stage key geometry.
For values, the shallow source is more strongly upweighted, indicating that lower-level content features remain useful when reconstructing the deep-layer memory content.
This qualitative pattern is consistent with the observation behind FusedKV.
However, the comparison in Table~\ref{tab:kv_sharing_comparison} shows that FusedKV still underperforms \method{} under the same 16B A3B MoE setting.
This indicates that the benefit does not come only from two-source reconstruction; distributing the source--reuse connections across layers, and thereby spreading the cross-layer gradient regularization over many participating source layers, is also important.

\FloatBarrier
\section{Efficiency and System Analysis}
\label{app:efficiency_system}

\subsection{KV Cache and Prefill Cost}
\label{app:kv_cache_prefill}

\method{} partitions the network into shallow, middle, and deep stages of approximately equal depth.
The shallow and middle stages materialize KV normally, while the deep stage reconstructs its KV from stride-aligned shallow--middle sources.
Therefore, no independent KV cache needs to be stored for the deep stage.
For an $L$-layer model with identical KV dimensionality across layers, the baseline stores KV for all $L$ layers, whereas \method{} stores KV only for roughly $2L/3$ source layers:
\begin{equation}
    \frac{\mathrm{KVCache}_{\method{}}}{\mathrm{KVCache}_{\mathrm{baseline}}}
    \approx \frac{2L/3}{L} = \frac{2}{3}.
\end{equation}
This gives an approximately one-third reduction in KV-cache memory during inference.

The same structure also enables prefill early completion.
In standard autoregressive inference, the prefill phase processes the full prompt through all layers to produce the KV cache and the hidden state needed to generate the first response token.
In \method{}, the purpose of prefill with respect to KV construction is completed once the shallow and middle stages have processed the full prompt, because all KV tensors needed by future decoding steps have already been materialized.
The deep stage does not need to process the entire prompt to create additional KV cache.
Instead, after the shallow and middle KV caches are available, we only forward the final prompt token through the deep stage, using the stored shallow--middle KV caches to compute attention and produce the first response token.

Thus, the prefill saving comes from avoiding full-sequence computation in the deep stage, while retaining the necessary computation for the last prompt token that initializes decoding.
As context length and model size increase, the full-prompt prefill computation becomes increasingly dominant, and the cost of forwarding only the final token through the deep stage becomes a smaller fraction of total prefill cost.
This explains the trend in Table~\ref{tab:efficiency_main}: the measured TTFT reduction grows at longer context lengths and approaches the theoretical one-third saving.

\subsection{Training FLOPs and Parameter Count}
\label{app:flops_params}

\method{} does not introduce additional training FLOPs in the way that many explicit cross-layer connectivity mechanisms do.
The shallow and middle stages are computed normally, and the deep-stage KV reconstruction only requires element-wise linear combinations of two cached source tensors.
This mixing cost is negligible compared with the matrix multiplications in attention and FFN layers.
Moreover, because the deep stage does not require independent K/V projections, an implementation can skip the corresponding K/V projection computation in reused layers, slightly reducing training FLOPs.
In our large-scale runs, the observed training throughput is essentially identical to the baseline.

We also reduce the activated parameter count by removing the K and V projection weights from reused layers.
For a GQA layer with hidden dimension $d_{\mathrm{model}}$, $n_{\mathrm{kv}}$ KV heads, and head dimension $d_h$, the K/V projection parameters per layer are
\begin{equation}
    P_{\mathrm{KV}}^{\mathrm{layer}}
    = 2 d_{\mathrm{model}} n_{\mathrm{kv}} d_h,
\end{equation}
where the factor of 2 accounts for $\mathbf{W}^K$ and $\mathbf{W}^V$.
If \method{} reuses KV in $L_{\mathrm{reuse}}$ layers, the removed activated parameters are
\begin{equation}
    \Delta P_{\mathrm{active}}
    = L_{\mathrm{reuse}} \cdot 2 d_{\mathrm{model}} n_{\mathrm{kv}} d_h.
\end{equation}
For the 16B A3B GQA MoE model, $d_{\mathrm{model}}{=}2048$, $n_{\mathrm{kv}}{=}8$, $d_h{=}128$, and $L_{\mathrm{reuse}}{=}9$.
This gives
\begin{equation}
    \Delta P_{\mathrm{active}}
    = 9 \times 2 \times 2048 \times 8 \times 128
    = 37.7\mathrm{M}.
\end{equation}
Relative to the 3B activated-parameter budget, this corresponds to approximately $1.3\%$ fewer activated parameters.
Thus, \method{} obtains its downstream gains while slightly reducing, rather than increasing, the active model capacity.

\subsection{Parallel Training Compatibility}
\label{app:parallel_training}

Pipeline parallelism (PP) is essential for training large Transformers whose parameters and activations cannot fit on a single device.
PP partitions the model along depth and assigns different layer blocks to different physical pipeline ranks.
Virtual pipeline parallelism (VPP) further splits each physical stage into smaller virtual chunks, typically at layer granularity, and interleaves their execution to reduce pipeline bubbles and improve hardware utilization.
For cross-layer KV sharing, the main systems concern is whether a reuse layer needs to fetch source KV tensors from a different PP rank; such accesses would introduce additional cross-stage communication and complicate activation lifetime management.

\method{} is designed to avoid this issue.
Because each deep layer reuses a stride-aligned shallow--middle source pair, the source and reuse layers can be co-located on the same physical PP rank under an appropriate VPP partition.
Table~\ref{tab:pp_vpp_layout} gives a balanced example for a 36-layer model with PP size 4.
Using zero-based layer indices, the network is divided into three 12-layer stages: layers 0--11, 12--23, and 24--35.
Each deep layer $i+24$ reconstructs KV from the stride-aligned pair $(i,i+12)$, and each triple $(i,i+12,i+24)$ is placed on the same physical PP rank.
Thus, cross-layer KV production and consumption remain local to one device.

\begin{table}[!htbp]
\centering
\small
\caption{A balanced PP/VPP-compatible layer placement for a 36-layer model with PP size 4. Each stride-aligned source--reuse triple is placed on the same physical PP rank.}
\label{tab:pp_vpp_layout}
\begin{tabular}{lcccc}
\toprule
\textbf{Depth band} & \textbf{PP rank 0} & \textbf{PP rank 1} & \textbf{PP rank 2} & \textbf{PP rank 3} \\
\midrule
Shallow sources & 0, 1, 2 & 3, 4, 5 & 6, 7, 8 & 9, 10, 11 \\
Middle sources  & 12, 13, 14 & 15, 16, 17 & 18, 19, 20 & 21, 22, 23 \\
Deep reuse      & 24, 25, 26 & 27, 28, 29 & 30, 31, 32 & 33, 34, 35 \\
\bottomrule
\end{tabular}
\end{table}

This layout preserves the natural forward order as three repeated PP passes over shallow, middle, and deep depth bands, while keeping the non-local KV dependency local within each physical rank.
With layer-granular VPP chunks, no additional inter-rank KV transfer is required for \method{} under this placement.
In this balanced setting, every PP rank owns exactly three shallow layers, three middle layers, and three deep reuse layers, so the per-rank layer count is identical.
For practical model depths that are not perfectly divisible by the number of PP ranks, the placement may introduce mild imbalance, but the final load remains close.
This locality property is important for scaling \method{} to large distributed pretraining runs without changing the communication pattern of the baseline PP/VPP system.

\end{document}